\documentclass[10pt,twocolumn,letterpaper]{article}

\usepackage[pagenumbers]{wacv}

\usepackage[table]{xcolor}
\usepackage{multirow}
\usepackage[accsupp]{axessibility}

\newlength\savewidth\newcommand\shline{\noalign{\global\savewidth\arrayrulewidth
		\global\arrayrulewidth 1pt}\hline\noalign{\global\arrayrulewidth\savewidth}}

\newcommand{\metricheader}[1]{\textbf{\scriptsize{\,#1\,}}}
\newcommand{\modelname}[1]{\textbf{\textcolor{gray}{\small{\,#1\,}}}}
\newcommand{\metricheadertwo}[1]{\textbf{\small{\,#1\,}}}

\definecolor{wacvblue}{rgb}{0.21,0.49,0.74}
\definecolor{tablegray}{RGB}{223, 242, 252}
\usepackage[pagebackref,breaklinks,colorlinks,allcolors=wacvblue]{hyperref}

\usepackage{caption}

\def\wacvPaperID{1752} %
\def\confName{WACV}
\def\confYear{2027}

\title{\textit{SurfScaff3D}: Bridging Geometric and Semantic Representations for LLM-based 3D Object Generation}

\author{Anay Loya\thanks{Equal contribution} \quad Rishi Gupta\footnotemark[1] \quad Arnav Samal \quad Anirban Chakraborty \\
Indian Institute of Science, Bengaluru, India\\
{\tt\small \{anayloya, rishig, anirban\}@iisc.ac.in, samalarnav@gmail.com}
}

\begin{document}

\twocolumn[{%
  \renewcommand\twocolumn[1][]{#1}%
  \maketitle
  \begin{center}
    \vspace{-6mm}
    Project Page: \href{https://vcl-iisc.github.io/SurfScaff3D/}{https://vcl-iisc.github.io/SurfScaff3D/}
  \end{center}
  \begin{center}
    \captionsetup{type=figure}
    \includegraphics[width=\textwidth]{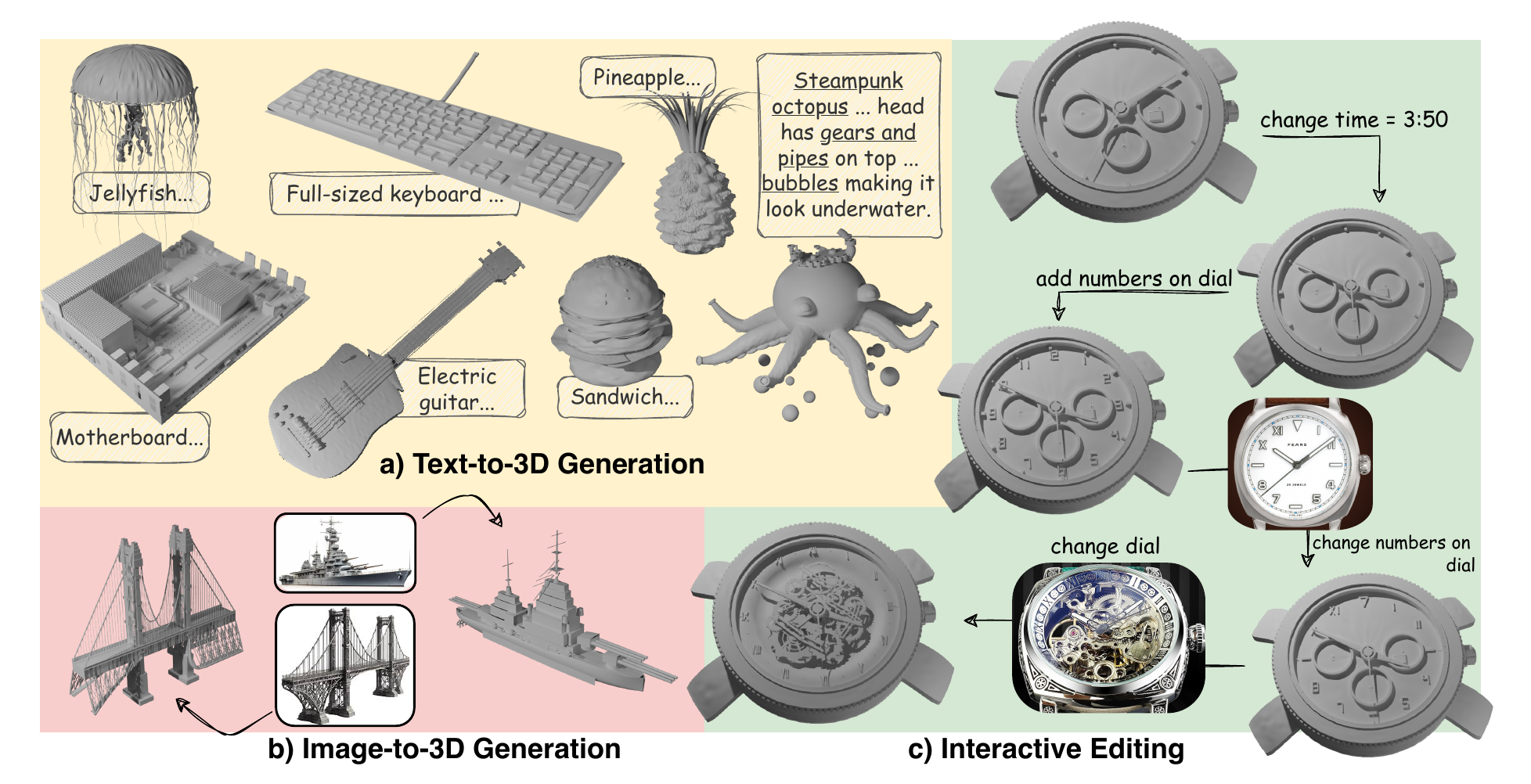}
    \captionof{figure}{\textbf{Capabilities of SurfScaff3D.} Our framework allows interactive text and image-to-3D generation, together with targeted editing. Built on our Hierarchical Patch Scaffold (HPS), SurfScaff3D combines interactive editability with high geometric flexibility, spanning both organic natural forms and intricate high-detail mechanical structures without imposing restrictive primitive priors.}
    \label{fig:teaser}
  \end{center}
  \vspace{1.0em}
}]

{\let\thefootnote\relax\footnotetext{$^*$Equal contribution, alphabetically ordered.}}

\begin{abstract}
Creating high-quality 3D content has traditionally required specialist expertise and experience with complex tools. The growing ecosystem of AI-assisted creativity demands accessible and intuitive 3D generation systems. Recent advances in Large Language Models (LLMs) offer promising directions, yet extending their generative capabilities to 3D remains fundamentally challenging because existing 3D representations are not naturally compatible with language-based reasoning. Current approaches either depend on expensive fine-tuning or impose restrictive structural priors that sacrifice expressiveness and exclude organic shapes. We argue that the mismatch between LLM-native reasoning and conventional 3D parameterizations is a core bottleneck of LLM-driven 3D generation. To address this, we introduce the Hierarchical Patch Scaffold (HPS), a representation designed around semantically meaningful, interpretable part-level abstractions that are naturally understood by both humans and LLMs, enabling expressive, editable, and open-domain 3D generation without imposing restrictive geometric priors or expensive fine-tuning.

Building upon HPS, we introduce \textbf{SurfScaff3D}, an interactive framework for text and image-to-3D generation. HPS inherits the flexibility of B-spline patches, while exposing an interface that leverages open-world capabilities of foundation models to reliably generate 3D. Its hierarchical part-level semantic labels enable targeted, natural-language edits at any level of granularity without re-running generation. Further, HPS offers explicit per-patch control over surface detail, enabling more expressive geometry capable of representing arbitrary objects. Extensive experiments against state-of-the-art LLM-based 3D generation methods demonstrate strong geometric fidelity, semantic alignment, and flexible editing across diverse object categories.

\end{abstract}
    
\section{Introduction}
\label{sec:intro}

3D content sits at the heart of modern visual computing, from industrial design and manufacturing \cite{you2026physgen} to game assets \cite{fukaya2025intelligent} and augmented reality overlays \cite{behravan2025augmented}. Yet creating high-quality 3D content remains a specialized task demanding expertise in professional tools, whose steep learning curves and laborious workflows require substantial manual effort even from professionals and remain wholly inaccessible to everyone else. Meanwhile, foundation models have democratized adjacent creative modalities such as text, image, audio and video generation, enabling non-experts to produce sophisticated content through natural prompts. Closing the analogous gap for 3D is an active research frontier. A practically useful system must be intuitive and interactive while remaining geometrically expressive, open-domain generalizable and directly editable. Existing LLM-based approaches do not achieve all of these at once, and we argue that this failure is representational, not architectural.

Researchers have pursued LLM-based 3D generation primarily along two directions which face a fundamental trade-off. Methods based on geometric \cite{wang2024llama-mesh, fang2025meshllm}, or parametric \cite{usama2026nurbgen} approaches offer flexible and expressive representations but require LLMs to predict low-level geometric parameters which are not natively aligned with LLMs and difficult to generate reliably. They often require fine-tuning, hence restricting them to their training distributions. Conversely, procedural \cite{zhou2026articraft, zhang2026shapecraft} or primitive-based \cite{yamada2025l3go} approaches expose higher-level abstractions that align naturally with language models and enable training-free generation, but these abstractions impose restrictive structural priors that limit geometric expressiveness and object diversity. Consequently, current methods typically trade representational expressiveness for open-domain generalization. We argue that this is a model-representation co-design mismatch, since existing representations were not designed jointly for geometric expressiveness and zero-shot LLM generation, forcing current methods to sacrifice one for the other.

This paper introduces the \textbf{Hierarchical Patch Scaffold (HPS)}, an intermediate representation explicitly designed to bridge this gap. HPS represents objects as a rooted semantic tree over a collection of B-spline patches. A chair, for instance, is structured hierarchically into semantic components such as the seat, backrest, and legs, each a coherent cluster of smooth patches. By decoupling high-level structural organization from underlying coordinate instantiation, HPS provides a geometrically flexible, interpretable, and scalable interface that also enables controllable smoothness at a granular, patch-level resolution.

HPS parameterizes each patch using surface points rather than the abstract control-point lattices of classical spline formulations. This surface-point interface is a key design choice underlying HPS, as surface points correspond directly to visible geometry. Similar observations were made by \cite{vinker2025sketchagent} in the related domain of sketch generation, and our ablations (\autoref{tab:hps_isolate}) further show that surface-point parameterization is more intuitive for LLMs than direct control-point prediction, as it better aligns with the geometric concepts expressed in natural language. The hierarchical semantic structure of HPS also enables intuitive and localized editing. Because semantic parts are explicitly mapped to geometric patch groups, edits can be targeted at arbitrary levels of granularity without regenerating the full object.

\begin{figure*}
    \centering
    \includegraphics[width=\linewidth]{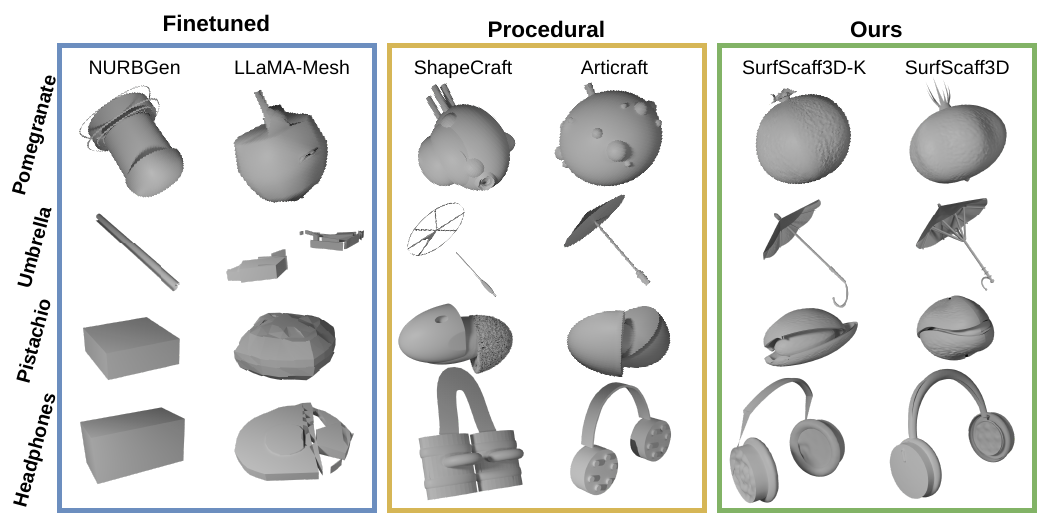}
    \caption{\textbf{Qualitative Comparison of SurfScaff3D.} We provide qualitative examples of SurfScaff3D against similar works, on the left we represent methods which allow greater geometric flexibility but require fine-tuning, in the middle we show methods which benefit from the zero-shot generalizability of modern LLMs but suffer from restrictive geometric priors. On the right we show SurfScaff3D using the open-source Kimi K2.6 \cite{moonshotai2026kimik26} and Claude Sonnet 4.6 \cite{anthropic2026claudeSonnet46SystemCard} as generation backbones, respectively. Each row corresponds to a single object prompt, all methods are generated using the identical prompts to ensure a fair qualitative comparison.}
    \label{fig:comparison}
\end{figure*}

Building upon HPS, we introduce \textbf{SurfScaff3D} (short for Surface Scaffold in 3D), a training-free 3D generation framework that enables programmatic construction of surface geometry that can be rendered into views for direct inspection of local geometry. The framework supports critic-driven iterative refinement for prompt alignment and global structural refinement along with interactive targeted editing. By grounding generation in executable programs with rendering-based feedback, SurfScaff3D avoids reliance on task-specific 3D supervision, generalizes across object categories, and naturally benefits from underlying improvements in foundation models. Extensive evaluations demonstrate that SurfScaff3D outperforms both procedural and fine-tuned baselines as well as a concurrent work, Articraft \cite{zhou2026articraft}, under comparable settings. Notably, unlike Articraft, which focuses on human evaluation, we additionally provide quantitative metrics on their setting, enabling comparisons against SurfScaff3D. Qualitative results in \autoref{fig:teaser} and \ref{fig:comparison} show that SurfScaff3D generates more organic yet structured shapes and supports increasingly complex and highly detailed objects.

\noindent Our contributions can be summarised as follows:
\begin{itemize} 
    
    \item \textbf{Hierarchical Patch Scaffold (HPS)}: We introduce a hierarchical B-spline patch-based surface representation that is geometrically flexible, interpretable, and scalable. HPS structures surfaces as a tree of patches and enables controllable smoothness at the patch level, allowing explicit control over local shape properties such as smoothness and organicity.

     \item \textbf{SurfScaff3D}: We propose an interactive training-free text and image-to-3D generation framework that leverages zero-shot multimodal reasoning in foundation models for open-domain 3D synthesis. %

    \item Comprehensive experiments across datasets spanning CAD-like, structural and organic objects demonstrate that SurfScaff3D outperforms both procedural and fine-tuned state-of-the-art LLM-based 3D generation baselines across generation and editing tasks.
\end{itemize}

\label{sec:related_works}

\section{Related Works}

\paragraph{Text/Image to 3D Generation.}
Diffusion and flow based methods span point-clouds such as Point-E~\cite{nichol2022pointe}, CAD-oriented representations such as BrepGen~\cite{xu2024brepgen} and CADCrafter~\cite{chen2025cadcrafter}, and implicit representations such as Shap-E~\cite{jun2023shap-e}, TRELLIS~\cite{xiang2025trellis} and TRELLIS.2~\cite{xiang2026native}. Multi-view diffusion priors such as MVDream \cite{shi2024mvdream} improve cross-view consistency for score-based text-to-3D generation. 

However, these methods require large-scale training and slow inference procedures. Moreover, opaque latent representations are often difficult to edit after generation, since modifying a localized part generally requires rerunning the full generative process. These limitations have motivated recent interest in leveraging LLMs for 3D generation.

\paragraph{Geometric Representations.}
A major challenge in LLM-driven 3D generation is the choice of representation exposed to them. LLaMA-Mesh \cite{wang2024llama-mesh} represents meshes as text sequences of vertices and faces. MeshLLM \cite{fang2025meshllm} further improves this paradigm by decomposing meshes into primitive tokens. More recently, ShapeLLM-Omni~\cite{wang2026shapellm} proposes a joint latent token space between VAEs and LLMs for native 3D understanding. However, mesh representations remain fundamentally discretized and organic geometry requires increasingly dense tessellation, limiting geometric expressiveness as shown in \autoref{fig:comparison}.
Moreover, these approaches require fine-tuning LLMs. Our work investigates whether the representation itself can naturally be aligned with the reasoning capabilities already present in them.

\paragraph{CAD-based Representations.} 

Another line of work adapts LLMs to structured CAD representations. CAD-LLaMA~\cite{li2025cadllama} formulates CAD construction histories as executable code sequences, while CAD-Code-Verify~\cite{al2025cadcodeverify} introduces iterative visual verification loops for CAD program refinement. BlenderLLM~\cite{du2024blenderllm} fine-tunes an LLM via self-improvement to generate executable Blender scripts. While these approaches benefit from the precision and editability of CAD representations, they remain tightly coupled to procedural CAD operations and engineering-oriented datasets.

\paragraph{Procedural and Primitive-Based Representations.}

A parallel line of work argues that modern LLMs already possess sufficient reasoning to support training-free 3D generation. L3GO~\cite{yamada2025l3go} demonstrates fine-tuning free generation through procedural abstractions that simplify the geometric reasoning problem for LLMs. Similarly, ShapeCraft~\cite{zhang2026shapecraft} employs LLM agents within Blender to iteratively construct and edit assets through a limited set of procedural operations. A concurrent work, Articraft~\cite{zhou2026articraft} extends this paradigm to articulated object generation with a single LLM agent that writes code against a domain-specific SDK. These approaches heavily rely on predefined procedural or CAD operators, which constrain the generation of organic geometry. Our work removes the fixed primitive assumptions entirely by representing geometry as hierarchically organized splines.

\paragraph{Smooth Parametric Representations.}
B-splines and NURBS have a long history in geometric modelling \cite{piegl2012nurbsbook}. Recently, SketchAgent~\cite{vinker2025sketchagent} employed splines for LLM-based sketch generation, while \cite{worchel2023differentiable} introduced differentiable rendering of parametric surfaces for inverse graphics. Closest to our work, NURBGen~\cite{usama2026nurbgen} fine-tunes LLMs to predict NURBS parameters for CAD model generation. However, because abstract control points are not natively understood by LLMs, NURBGen relies on fine-tuning, limiting itself to domain-specific CAD objects. Our HPS is co-designed from the outset as an LLM-native interface for generation and editing rather than a mere fitting target.

\paragraph{3D Editing.}
Recent work has adapted diffusion priors for fine-grained 3D editing. SDEdit~\cite{meng2022sdedit} and DiffEditor~\cite{mou2024diffeditor} introduced guided image editing through diffusion. GeoDiffusion~\cite{mueller2025geodiffusion} extends this to use 3D geometric priors for image-level editing, and MVDrag3D~\cite{chen2024mvdrag3d} extends drag-based editing directly to 3D assets using multi-view guidance. Prox-E~\cite{sella2026proxe} enables training-free 3D editing through an explicit primitive-based shape abstraction. These approaches however, cannot guarantee identity preservation. On the other hand, HPS relies on explicit hierarchical representation that enables targeted editing of named parts without latent-space manipulation or full-object regeneration. We provide a more detailed characteristic comparison of different editing methods in Appendix D.

\section{Preliminary: B-Splines}
\label{sec:b-splines}

B-Splines form the backbone of HPS. We briefly review the fundamentals of 1D B-spline curves and their tensor-product extensions to higher dimensions.

\noindent\textbf{1D Curves. }
A B-spline curve, $C(u) = \sum P_i N_{i}^{(p)}(u)$, is defined by its control points $P_i$ and degree-$p$ basis functions $N_{i}^{(p)}(u)$ over the parameter domain $u \in [0,1]$.

 The basis functions are defined over a non-decreasing knot vector $U = \{u_0, \dots, u_m\}$, whose entries determine the intervals over which the polynomial pieces are active. In contrast to a single high-degree B\'ezier curve, B-splines are constructed from piecewise polynomials at knots, enabling complex shapes while preserving local control. The Cox-de Boor recursion defines the basis functions: 
\begin{equation}
\begin{split}
    N_{i}^{(0)}= \begin{cases}
    1 & \text{if } u_i\le u < u_{i+1} \\
    0 & \text{otherwise}\phantom{phantom phano,.}
\end{cases} 
\\
N_{i}^{(p)} = \frac{u-u_i}{u_{i+p}-u_i} N_{i}^{(p-1)}(u) + \phantom{phantom}\\ 
\phantom{more phatnom}
\frac{u_{i+p+1}-u}{u_{i+p+1}-u_{i+1}} N_{i+1}^{(p-1)}(u)
\end{split}
\end{equation}
where terms with zero denominators are conventionally defined to be zero \cite{piegl2012nurbsbook}.

\noindent\textbf{Tensor Product B-Splines. }
The same construction extends naturally from curves to higher-dimensional spline mappings through tensor products: $S = S_1\otimes S_2 \otimes \dots \otimes S_d$, where $S_k$ is the univariate spline space of the $k^{\text{th}}$ parameter direction.

Let $(u_1,\dots,u_d)$ denote coordinates in a $d$-dimensional parameter domain. A tensor-product B-spline is defined as:
\begin{equation}
    C(u_1, \dots, u_d) = \sum_{i_1 = 0}^{n_1} \cdots \sum_{i_d = 0}^{n_d} \prod_{k=1}^d N_{i_k}^{(p_k)}(u_k) P_{i_1 \dots i_d}
\end{equation}

where $\{N_{i_k}^{(p_k)}\}_{i_k=0}^{n_k}$ is the set of basis functions associated with the $k^{\text{th}}$ parameter direction and $P_{i_1 \dots i_d}$ are arranged in a control lattice. In this work, we fit these 2D tensor-product B-splines to the generated surface points to instantiate individual 3D patches.

\section{Methodology}
\begin{figure*}[t]
    \centering
    \includegraphics[width=\linewidth]{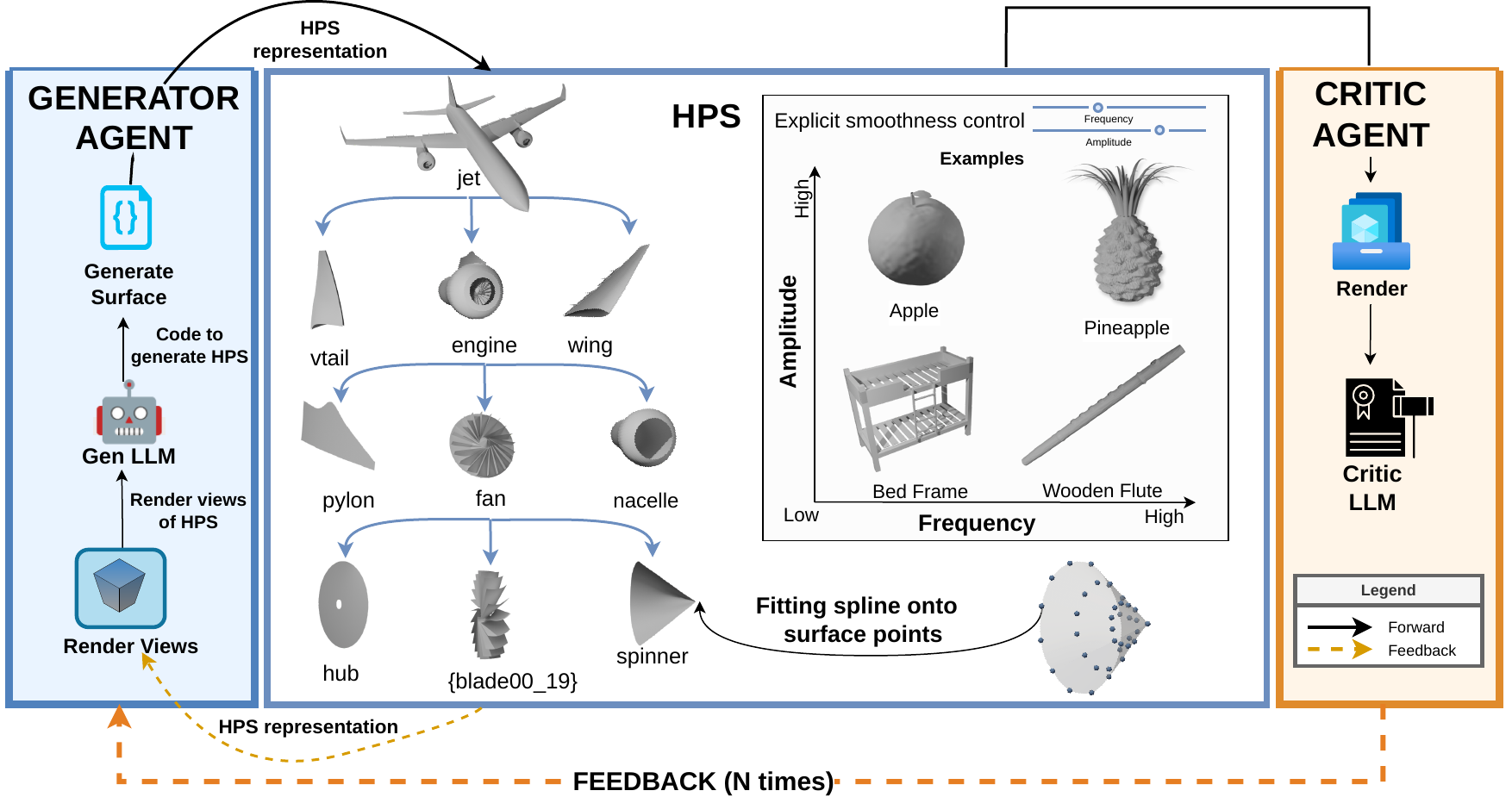}
    \caption{\textbf{Overview of SurfScaff3D}. SurfScaff3D iteratively constructs an HPS representation by programmatically generating surface geometry and verifying its local geometry through rendered views. HPS organizes geometry as a hierarchical semantic tree of B-spline surface patches with explicit patch-level control over surface variation, enabling both structured mechanical objects and organic shapes. The examples illustrate how frequency and amplitude help in modelling diverse objects. A critic evaluates rendered outputs and provides targeted feedback, forming a closed-loop refinement process that progressively improves prompt alignment and geometric fidelity. The critic loop executes for a fixed number of N iterations unless an early stopping criterion is met.}
    \label{fig:model}
\end{figure*}

\subsection{Hierarchical Patch Scaffold}

We introduce the Hierarchical Patch Scaffold (HPS), a geometric representation explicitly co-designed to bridge the gap between high-level language reasoning and continuous, editable 3D shapes. Formally, HPS decomposes a 3D asset into a tuple $(\mathcal{T}, \Pi)$, where $\mathcal{T}$ represents a global semantic hierarchy tree and $\Pi$ denotes the set of surface patches. Each individual patch $f \in \Pi$ is parameterized as $(\mathcal{P}, s_f, \omega_f, l)$. Here, $\mathcal{P}$ represents the surface points, the parameters $s_f$ and $\omega_f$ explicitly govern the local geometric texture, finally, $l$ represents part-level annotations that provide semantic grounding to the overall representation. We validate the impact of our HPS representation in \autoref{tab:hps_isolate}.

\paragraph{Surface-Point Interface.}
The core component, $\mathcal{P} \in \mathbb{R}^{n_u \times n_v \times 3}$, represents a structured grid of sparse 3D surface points generated via tool-assisted code execution. 2D B-splines are fitted onto these points, abstracting spline parameterization away from the LLM. Because LLMs are pre-trained on massive text and vision corpora that describe visible surface geometry, programming explicit surface points serves as a more natural interface than directly predicting spline parameters. We also validate this in \autoref{tab:hps_isolate} by replacing surface points with B-Spline control points.%

\paragraph{Hierarchical Annotations.} The structural backbone of the HPS is formalized as a tree $\mathcal{T} = (\mathcal{V}, \mathcal{E})$. To bridge high-level semantic hierarchy with the discrete geometry, each node $v \in \mathcal{V}$ represents a semantic concept linked to a subset of surface patches via an association function $\phi: \mathcal{V} \rightarrow 2^{\Pi}$. %

\paragraph{Patch-Level Surface Variation.}

\begin{figure}
    \centering
    \includegraphics[width=\linewidth]{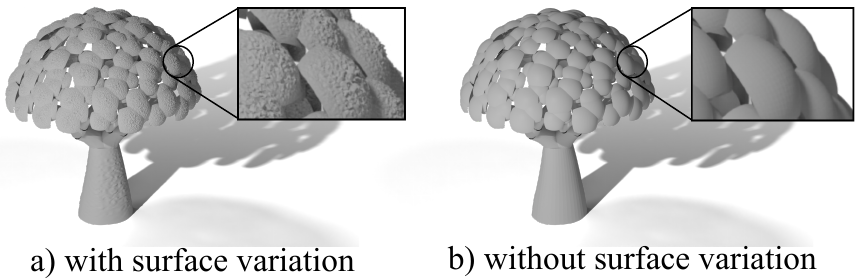}
    \caption{\textbf{Surface Variation}. An organic broccoli model generated by SurfScaff3D, demonstrating the expressive capacity of the HPS smoothness parameters. By modulating the local smoothness parameters $s_f$ and $\omega_f$, the model synthesizes complex, high-frequency organic geometry without losing geometric coherence.}
    \label{fig:surface variation}
\end{figure}

To ensure organic generation, we equip each face $f\in\Pi$ in our asset with the tuple $(s_f, \omega_f)$, where $s_f \in [0.05,1]$ represents the smoothness of the surface, hence its reciprocal depicts the amplitude of surface variations and  $\omega_f \in [1, 10]$ represent their frequency. 
In \autoref{fig:model} we demonstrate how modulating these parameters benefits diverse real-world objects, allowing explicit, fine-grained, manual control over the generated surface. We show the impact of surface variation in \autoref{fig:surface variation}.

First, the initial point grid of size $n_u \times n_v$ is upsampled by a factor of $\omega_f$, yielding a denser grid of size $(\omega_f\cdot n_u) \times (\omega_f \cdot n_v)$. We then model the surface variation on each point as:
\begin{equation}
p := p+\frac{d\cdot\sigma}{s_f}\mathcal{N}(0,\mathcal{I})
\label{eq:surface_variation}
\end{equation}

Here, $\mathcal{N}$ represents Gaussian noise, $\sigma$ is our base noise scaling parameter, and $d=\min\left(\alpha, \min_{e\in f_{\mathrm{edge}}}\|p-e\|_2\right)$ represents the Euclidean distance of the upsampled point $p$ to the nearest point on the patch boundary $f_{\mathrm{edge}}$, clipped at $\alpha$.
By weighting the noise amplitude by $d$, we ensure that the variation gradually tapers to zero as we move towards the boundaries of a face. This enforces that irregularities are suppressed near structural edges, maintaining consistency across adjacent patches.

\subsection{SurfScaff3D}

Building upon HPS, we introduce \emph{SurfScaff3D}: an interactive, interpretable, and training-free framework for 3D generation. Our system is designed around the observation that effective 3D generation using LLMs requires representations that align naturally with their spatial reasoning.

Therefore, SurfScaff3D couples HPS with an iterative agentic generation pipeline. Our framework employs a dual-agent architecture consisting of a \textit{Generator Agent} and a \textit{Critic Agent}. The generator incrementally constructs geometry through programmatic surface synthesis and iterative visual verification, while the critic evaluates structural fidelity, prompt alignment, and semantic consistency to guide self-correction. Together, these agents form a closed refinement loop that progressively improves the generated geometry and prompt alignment, as illustrated in \autoref{fig:model}.

We employ Claude Sonnet 4.6~\cite{anthropic2026claudeSonnet46SystemCard} as the primary backbone of our generator agent, alongside two
open-source alternatives: Kimi K2.6~\cite{moonshotai2026kimik26}, and Qwen3.6-35B-A3B~\cite{qwen2026qwen35b} as a lightweight option. We name these pipelines SurfScaff3D-K and SurfScaff3D-Q, respectively. For all variants Gemini 2.5 Pro~\cite{gemini25} was used as the critic agent. We report numbers using other backbones in Appendix E, along with the prompt templates in Appendix A.1.
 \begin{itemize}
     
\item \textbf{Generator Agent.}
The generator agent receives task instructions, in-context examples, and optionally, reference images. It uses a tool-calling framework equipped with two primary utilities: \texttt{generate\_surface} and \texttt{render\_views}. Rather than relying on its own spatial reasoning to directly predict surface points, the generator agent synthesizes Python code utilizing standard mathematical libraries to programmatically construct the structural geometry. The \texttt{generate\_surface} tool executes this script in an isolated sandbox, returning either the resulting surface patches or the execution traces to the model. The \texttt{render\_views} tool allows the agent to visualize local patches to verify geometry by rendering multiple views. The generator can invoke these tools iteratively, enabling adjustments and visual verification before finalizing the output.

\item \textbf{Critic Agent. }
To enforce geometric fidelity and prompt alignment, we employ a critic agent that evaluates intermediate HPS representations. The critic receives multi-view renderings along with the original text prompt and any conditioning images. While local view consistency is handled by the rendering pipeline available to the generator agent, the critic is responsible for assessing global attributes such as overall geometry, prompt alignment, and naturalness of the asset. The critic scores the generation on a continuous scale within $[0, 1]$ and identifies up to three specific geometric or semantic discrepancies. For each identified issue, the critic produces an actionable response comprising the target part, the observed problem, and a prescriptive fix. This refinement process employs an early-stopping mechanism based on a target score threshold and is bounded to a fixed number of iterations, ultimately selecting the highest-scoring generation as judged by the critic agent.
\begin{figure}
    \centering
    \includegraphics[width=\linewidth]{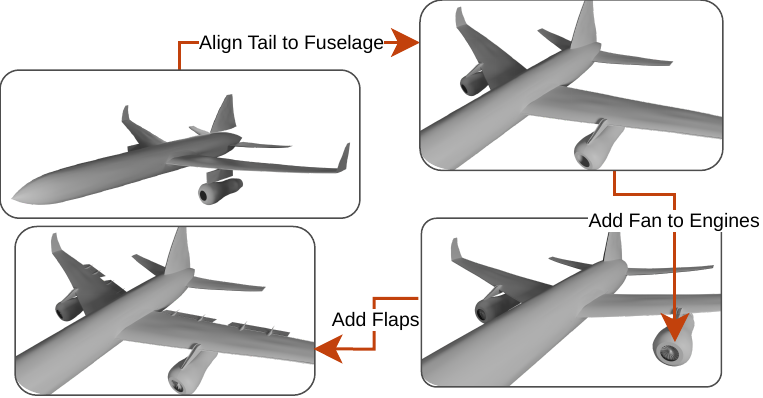}
    
    \caption{\textbf{Human-as-the-Critic Refinement.} SurfScaff3D allows integration of the user as a real-time critic within the generative loop. By providing natural language feedback, the human critic can drive targeted and personalized refinements.}
    \label{fig:human-critic}
\end{figure}
\end{itemize}

\begin{table*}[!ht]
    \centering
    \small
    
    \setlength\tabcolsep{1.5pt}
    \begin{tabular}{
        lccccc|ccccc|ccccc|c
    }
    
    \multicolumn{1}{c}{} & \multicolumn{5}{c}{\textbf{\textcolor{gray}{ShapeNetCore}}} & \multicolumn{5}{c}{\textbf{\textcolor{gray}{Toys4K}}} &
    \multicolumn{5}{c}{\textbf{\textcolor{gray}{OmniObject3D}}} &  \\

    \multicolumn{1}{c}{} & 
    \multicolumn{3}{c}{Prompt Align} & \multicolumn{2}{c}{3D Qual} &
    \multicolumn{3}{c}{Prompt Align} & \multicolumn{2}{c}{3D Qual} &
    \multicolumn{3}{c}{Prompt Align} & \multicolumn{2}{c}{3D Qual}
    & Failure
    \\

     \cmidrule(lr){2-4} \cmidrule(lr){5-6} \cmidrule(lr){7-9} \cmidrule(lr){10-11} \cmidrule(lr){12-14} \cmidrule(lr){15-16} \noalign{\vskip -0.5ex}

    \textbf{Text-to-3D Model} 
    & \metricheader{CLIP$\uparrow$}  & \metricheader{ULIP$\uparrow$} & \metricheader{VQA$\uparrow$}  & \metricheader{FPD$\downarrow$} & \metricheader{Elo$\uparrow$}
    & \metricheader{CLIP$\uparrow$} & \metricheader{ULIP$\uparrow$} & \metricheader{VQA$\uparrow$}  & \textbf{\metricheader{FPD$\downarrow$}} & \textbf{\metricheader{Elo$\uparrow$}} 
    & \metricheader{CLIP$\uparrow$} & \metricheader{ULIP$\uparrow$} & \metricheader{VQA$\uparrow$}  & \metricheader{FPD$\downarrow$} & \metricheader{Elo$\uparrow$}
    & (\%)
    
    \\
                    
    \shline \multicolumn{1}{c}{} & \multicolumn{5}{c|}{\vspace*{-8pt}} & \multicolumn{5}{c|}{} & \multicolumn{5}{c|}{} &\\

    BlenderLLM \cite{du2024blenderllm} 
    & 24.9 & 21.9 & 38.3 & 930 & \phantom{0}763
    & 24.2 & 20.6 & 35.3 & \phantom{0}397 & \phantom{0}670 
    & 25.2 & 19.1 & 41.9 & 991 & \phantom{0}780
    &36.4
    \\

    LLaMA-Mesh \cite{wang2024llama-mesh} \;\;\;\;\;\;\;\;\;\;\;
    & 23.4 & 16.6 & 24.4 & 305 & \phantom{0}696
    & 23.2 & 13.2 & 21.8 & \phantom{0}298 & \phantom{0}648 
    & 23.8 & 15.8 & 37.9 & 298 & \phantom{0}646
    & \phantom{3}0.9
    \\
    
    NURBGen \cite{usama2026nurbgen}
    & 23.6 & \phantom{0}7.3 & 23.0 & 736 & \phantom{0}716
    & 23.4 & \phantom{0}6.6 & 19.6 & 1198 & \phantom{0}759 
    & 24.1 & \phantom{0}7.6 & 23.7 & 706 & \phantom{0}660
    & \phantom{0}0.9
    \\

    \hline

    ShapeCraft \cite{zhang2026shapecraft}
    & 24.6 & 20.6 & 38.8 & 378 & \phantom{0}910
    & 24.9 & 20.8 & 39.6 & \phantom{0}279 & 1030 
    & 25.1 & 20.1 & 42.0 & 251 & \phantom{0}922
    & \phantom{0}1.3
    \\
    
    Articraft \cite{zhou2026articraft}
    & 28.0 & 23.3 & 58.8 & 258 & 1258
    & 27.8 & 24.4 & 59.0 & \phantom{0}200 & 1192
    & 27.8 & 24.1 & 60.6 & 234 & 1189
    & \phantom{0}3.0
    \\
    
    \hline
    \rowcolor{tablegray}
    
    \textbf{SurfScaff3D-Q}
& 26.8 & 20.8 & 51.6 & 397 & 1005
& 26.9  & 23.4  & 59.6  & \phantom{0}394   & 1149
& 28.1  & 22.8  & 60.5  & 384   & 1161
& \phantom{0}\textbf{0.0}
\\
    \rowcolor{tablegray}

    \textbf{SurfScaff3D-K}
    & \underline{28.9} & \underline{23.4} & \textbf{66.9} & \underline{195} & \underline{1301} 
    & \textbf{29.2} & \underline{25.8} & \textbf{61.7} & \phantom{0}\textbf{135} & \textbf{1292} 
    & \textbf{28.8} & \underline{24.8} & \textbf{65.1} & \underline{211} & \underline{1284}
    & \phantom{0}\textbf{0.0}
    \\

    \rowcolor{tablegray}

    \textbf{SurfScaff3D}
    & \textbf{29.2} & \textbf{24.1} & \underline{66.6} & \textbf{186} & \textbf{1353}
    & \underline{28.9} & \textbf{26.2} & \textbf{61.7} & \phantom{0}\textbf{135} & \underline{1260}
    & \underline{28.5} & \textbf{25.5} & \underline{64.8} & \textbf{206} & \textbf{1356}
    & \phantom{0}\textbf{0.0}
    \\

    \specialrule{1pt}{0pt}{0pt}

    \end{tabular}
    \caption{\textbf{Evaluation of Text-to-3D Generation.} We evaluate the prompt fidelity using CLIP, ULIP scores, and VQA pass rate and 3D quality using FPD~\cite{shu2019fpd} and an Elo scoring based on GPT-5.4-mini \cite{openai2026gpt54} preferences (GPT represents an independent LLM family  which was not a part of any of the evaluated pipelines). We evaluate generation across three 3D datasets on captions from MARVEL-40M+ \cite{sinha2025marvel}. The zero-failure rate of SurfScaff3D across different backbone variants demonstrates the reliability of HPS as an intermediate representation.}
    \label{tab:gen_main}
\end{table*}

\noindent\textbf{Human-in-the-Loop Refinement.} 
Our method naturally accommodates targeted human-in-the-loop refinement, allowing users finer control over generation. ~\autoref{fig:human-critic} demonstrates an example of this manual refinement.

\section{Experiments}
\label{sec:Exp}

\noindent\textbf{Generation. }
We evaluate the text-to-3D generation capabilities of SurfScaff3D using prompts from MARVEL-40M+ \cite{sinha2025marvel}. We select test objects from three distinct datasets: ShapeNet-Core~\cite{chang2015shapenet} for clean CAD-style objects, Toys4K~\cite{stojanov2021using} for toy-like assets, and OmniObject3D \cite{wu2023omniobject3d} for real-world, organic objects. We report quantitative results in \autoref{tab:gen_main}. Since Articraft allows for any LLM backbone, we instantiate it with Kimi K2.6 to establish an identical foundation for evaluation against SurfScaff3D-K.

\begin{figure}
    \centering
    \includegraphics[width=\linewidth]{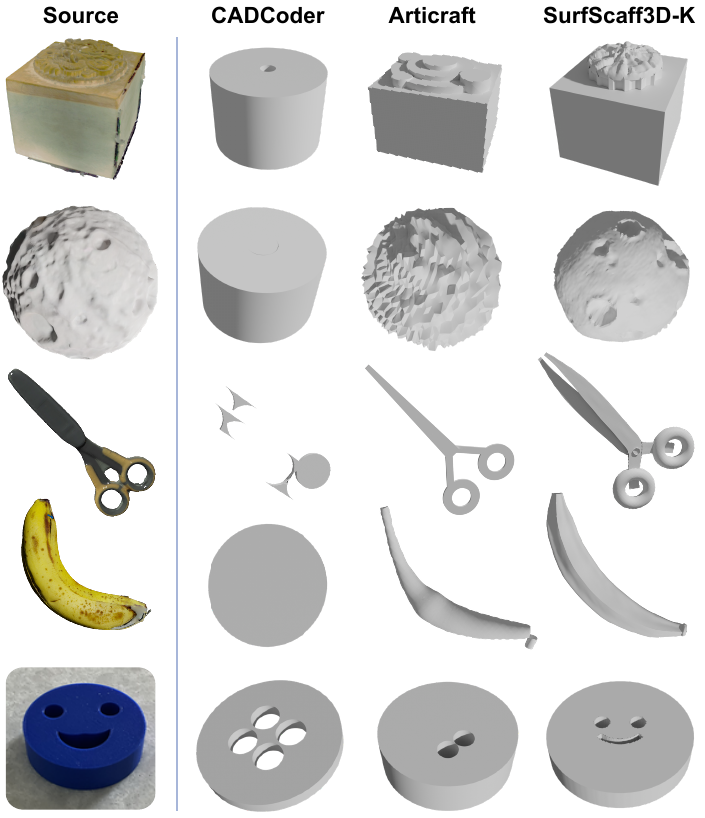}
    \caption{\textbf{Qualitative Image-to-3D Comparison.} We present qualitative comparisons of various organic, synthetic and CAD objects. The source image is provided without any textual description of the object. HPS allows greater flexibility in geometry.}
    \label{fig:comparison-image-to-3d}
\end{figure}

We assess text alignment via CLIP~\cite{radford2021clip}, ULIP~\cite{xue2023ulip} scores, and VQA pass rate. Geometric quality is measured via Fr\`echet Point Cloud Distance (FPD) \cite{shu2019fpd} and an Elo rating computed using the Bradley-Terry model~\cite{bradley1952rank}. Pairwise preferences for the Elo rating are judged by GPT-5.4-mini \cite{openai2026gpt54}, ensuring the LLM-family used for evaluations is independent of the pipelines evaluated.
We prompt the judge to make a decision by taking into account structural coherence, detail quality, and surface appropriateness. We further evaluate image-to-3D generation capabilities of SurfScaff3D on OmniObject3D and on the real-photograph test set released with CAD-Coder~\cite{doris2026cadcoder}, which consists of photographs of 50 3D-printed objects from the DeepCAD~\cite{wu2021deepcad} test split. We use one photograph per object. We report Chamfer distance and F-score at 0.05 threshold with the ground truth mesh along with CLIP score and LPIPS~\cite{zhang2018lpips}, as shown in \autoref{tab:img23d}. Since the code for several methods~\cite{chen2025cadcrafter, li2025cadllama, al2025cadcodeverify, yamada2025l3go} is not publicly available, we have not included them in our comparisons. Qualitative comparisons for Image-to-3D generation for some objects from OmniObject3D and DeepCAD are shown in \autoref{fig:comparison-image-to-3d}.

\begin{table}[!ht]
{\color{black}
    \centering
    \small
    \setlength\tabcolsep{1.5pt}
    \begin{tabular}{
        ll cccc
    }

    \textbf{Dataset} & \textbf{Method}
    & \metricheader{Chamfer$\downarrow$} & \metricheader{F@0.05$\uparrow$} & \metricheader{CLIP-I$\uparrow$} & \metricheader{LPIPS$\downarrow$}
    \\

    \shline \multicolumn{2}{c}{} & \multicolumn{4}{c}{\vspace*{-8pt}}\\

    \multirow{3}{*}{\metricheader{\textcolor{gray}{OmniObject3D}}}
    & CADCoder \cite{doris2026cadcoder}
    & 0.154 & 0.424 & 66.1 & 0.402
    \\
    & Articraft \cite{zhou2026articraft}
    & \underline{0.096} & \underline{0.596} & \underline{67.7} & \underline{0.295}
    \\
    \rowcolor{tablegray}
    \cellcolor{white}& \textbf{SurfScaff3D-K}
    & \textbf{0.084} & \textbf{0.709} & \textbf{71.0} & \textbf{0.269}
    \\

    \specialrule{0.4pt}{1pt}{1pt}

    \multirow{3}{*}{\metricheader{\textcolor{gray}{DeepCAD~\cite{wu2021deepcad}}}}
    & CADCoder \cite{doris2026cadcoder}
    & 0.140 & 0.464 & 71.0 & 0.400
    \\
    & Articraft \cite{zhou2026articraft}
    & \underline{0.115} & \underline{0.570} & \underline{73.3} & \underline{0.347}
    \\
    \rowcolor{tablegray}
    \cellcolor{white}& \textbf{SurfScaff3D-K}
    & \textbf{0.094} & \textbf{0.642} & \textbf{75.8} & \textbf{0.319}
    \\

    \specialrule{1pt}{0pt}{0pt}

    \end{tabular}
    \caption{\textbf{Image-to-3D.} We evaluate image-conditioned 3D
    generation on OmniObject3D and DeepCAD, reporting geometric accuracy (Chamfer, F@0.05), conditioning fidelity (CLIP-I), and novel-view reconstruction quality (LPIPS).}
    \label{tab:img23d}
}
\end{table}

While SurfScaff3D demonstrates consistent improvements over the clean CAD-like geometry of ShapeNet-Core, its advantage becomes more pronounced on complex objects from OmniObject3D and Toys4K. This highlights the geometric expressiveness of HPS. Additionally, SurfScaff3D's low compilation failure rate highlights the robustness of HPS as an LLM-to-geometry interface. More qualitative results are provided in \autoref{fig:teaser} and Appendix B. We observe a mean per-object cost of \$0.55 (Qwen), \$0.93 (Kimi), and \$1.58
(Claude) over 5 iterations. Running fewer iterations can further reduce this cost.

\noindent\textbf{Editing. }
Following Prox-E~\cite{sella2026proxe}, we evaluate text-driven 3D editing using instructions from ShapeTalk~\cite{achlioptas2023shapetalk}. We assess the edited outputs across CLIP, ULIP, and VQA pass rate and FPD. We report text-to-3D editing results in \autoref{tab:edit}. SurfScaff3D maintains a 0\% compilation failure rate during editing, whereas the baselines degrade significantly. This robustness demonstrates that HPS inherently serves as a stable, editable geometric interface for LLMs.  

\begin{table}[!ht]
    \centering
    \small
    
    \setlength\tabcolsep{1.5pt}
    \begin{tabular}{
        lccccc
    }

    \textbf{Model} 
    & \metricheader{CLIP$\uparrow$}  & \metricheader{ULIP$\uparrow$} & \metricheader{VQA$\uparrow$}  & \metricheader{FPD$\downarrow$} & \metricheader{Fail\%}
    \\
                    
    \shline \multicolumn{1}{c}{} & \multicolumn{5}{c}{\vspace*{-8pt}}\\

    LLaMA-Mesh \cite{wang2024llama-mesh} \;\;\;\;\;\;\;\;\;
    & \underline{26.6} & 21.0 & 32.4 & \underline{740} & 62.2 

    \\
    NURBGen \cite{usama2026nurbgen}
    & 24.9 & \phantom{0}9.4 & 42.6 & 786 & \phantom{0}\underline{1.1}
    \\

    ShapeCraft \cite{zhang2026shapecraft}
    & 25.9 & \underline{21.1} & \underline{59.7} & 909 & 31.1
    \\ 
    
    \rowcolor{tablegray}

    \textbf{SurfScaff3D-K}
    & \textbf{29.5} & \textbf{25.6} & \textbf{75.6} & \textbf{426} & \phantom{0}\textbf{0.0}
    \\
    
    \specialrule{1pt}{0pt}{0pt}

    \end{tabular}
    \caption{\textbf{Editing.} We evaluate the editing capability of various LLM-based text-to-3D models. We employ CLIP, ULIP and VQA scores along with FPD to evaluate the quality of the edited 3D assets and utilise edit prompts from ShapeTalk~\cite{achlioptas2023shapetalk}.}
    \label{tab:edit}
\end{table}

\subsection{Ablations}
\label{subsec:abl}
We evaluate the impact of some key components from our framework, motivating HPS and its design choices.

\begin{figure}[t]
    \centering
    \begin{minipage}{0.47\linewidth}
        \centering
        \includegraphics[width=\linewidth]{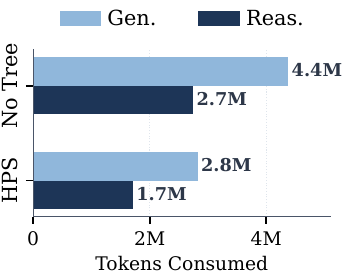}
        \caption{HPS allows more intuitive editing, consuming fewer tokens.}
        \label{fig:hps_token}
    \end{minipage}
    \hfill
    \begin{minipage}{0.47\linewidth}
        \centering
        \includegraphics[width=\linewidth]{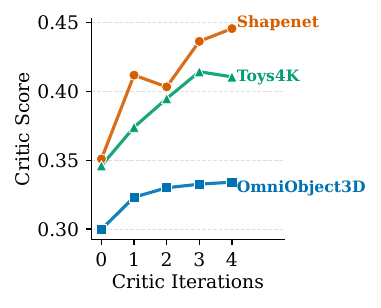}
        \caption{Critic score plateaus for SurfScaff3D (Claude Sonnet 4.6) after 2-3 iterations.}
        \label{fig:critic}
    \end{minipage}

\end{figure}

\noindent\textbf{Evaluating Impact of HPS. }
To quantify impact of HPS in our pipeline and differentiate it from the capability of the LLM alone we compare it against the following settings: 
\begin{itemize}
    \item \textbf{Point Cloud.} The agent produces a simple point cloud without additional structure, motivating the benefit of the overall HPS structure.
    \item \textbf{B-Spline Control Points.} This setting replaces surface points with control points, the more popular choice among parametric methods.
\end{itemize}
This experiment validates that HPS is indeed providing benefit and is more than just a point cloud or a set of B-Splines. We disable critic iterations and equip each method with correspondingly parameterized in-context examples for the same set of objects. As shown in \autoref{tab:hps_isolate}, we conduct this evaluation on our open-source backbones Kimi K2.6 and Qwen3.6-35B-A3B on OmniObject3D. Here, Elo scores are computed within a column, hence they should not be compared across columns. \autoref{fig:hps_token} motivates the benefit of the tree structure of HPS for editing. %

\noindent\textbf{Effect of Surface Variation. }
We evaluate the surface variation mechanism in \autoref{tab:sv_ablation}. Each generated object is exported with surface variation (HPS, our full method) and without it (w/o SV). GPT-5.4-mini picks the better of the two, using the same pairwise protocol as for the Elo scores. We report the number of objects for which each export is preferred.

\begin{table}[ht!]
    \centering
    \small                                 
    \begin{tabular}{lcc|cc}
        & \multicolumn{2}{c}{\modelname{Qwen3.6}} & \multicolumn{2}{c}{\modelname{KimiK2.6}} \\
        \metricheadertwo{Dataset} & \metricheadertwo{HPS} & \metricheadertwo{w/o SV} & \metricheadertwo{HPS} & \metricheadertwo{w/o SV}\\
        \shline \multicolumn{1}{c}{} & \multicolumn{2}{c|}{\vspace*{-8pt}} & \multicolumn{2}{c}{}\\
        
        ShapeNetCore  & 30 & 25 & 31 & 24 \\
        Toys4K &  64 & 41 & 60 & 45\\
        OmniObject3D & 46 & 25 & 41 & 30 \\
        \bottomrule
    \end{tabular}
    
     \caption{\textbf{Impact of Surface Variation.} Pairwise preference (GPT-5.4-mini~\cite{openai2026gpt54}, same protocol as the Elo scores in \autoref{tab:gen_main}) between the HPS output and the same output exported without surface variation (w/o SV), per dataset, across two open-source backbones. For each backbone, the two columns add up to the number of objects in the dataset (55, 105 and 71). Higher is better $\uparrow$.}
    \label{tab:sv_ablation}

\end{table}

\noindent\textbf{Effect of Hierarchical Labels on Editing. }
As shown in \autoref{fig:hps_token}, removing semantic labels substantially increases both the reasoning and generation token consumption for editing. This suggests that the explicit semantic structure enforced by HPS enables the LLM to localize edits more efficiently and naturally.

\noindent\textbf{Saturation of Iterative Critic Refinement. }
As shown in \autoref{fig:critic}, we evaluate how the critic score changes across refinement iterations. We observe that the critic score plateaus after 2-3 refinement iterations and offers diminishing improvements per additional iteration.

\begin{table}[ht!]
    \centering
    \small                                  
    \begin{tabular}{lcc|cc}
        & \multicolumn{2}{c}{\modelname{Qwen3.6 35B-A3B}} & \multicolumn{2}{c}{\modelname{KimiK2.6 1T-A32B}} \\
        \textbf{\small{Setting}} & \metricheadertwo{FPD$\downarrow$} & \metricheadertwo{Elo Score$\uparrow$} & \metricheadertwo{FPD$\downarrow$} & \metricheadertwo{Elo Score$\uparrow$}\\
        \shline \multicolumn{1}{c}{} & \multicolumn{2}{c|}{\vspace*{-8pt}} & \multicolumn{2}{c}{}\\
        
        Cloud   & 929 & 936 & 628 & 863 \\
        Control & 1192 & 960 & 745 & 930 \\
        \textbf{HPS} & \textbf{571} & \textbf{1103} & \textbf{391} & \textbf{1206} \\
        \bottomrule
    \end{tabular}
    \caption{\textbf{Isolating Impact of HPS.} In Cloud, the agent outputs a point cloud without additional structure. In Control, each patch is specified by control points instead of surface points. HPS is far more effective than both. All settings are evaluated without critic iterations and with the same set of in-context objects, to isolate the impact of the representations.}

    \label{tab:hps_isolate}
\end{table}

\section{Conclusion}
 
We introduced HPS, a representation that organises continuous B-spline patches into a rooted semantic tree, and SurfScaff3D, a framework that bridges language reasoning with 3D shape generation. By decoupling semantic planning from coordinate generation, HPS serves as an interpretable language-to-geometry interface outperforming baselines across datasets.

\paragraph{Limitations.}
HPS does not explicitly enforce watertightness or self-intersection avoidance and therefore provides no formal guarantees for these properties. Its fidelity is also bounded by prompt descriptiveness and the spatial reasoning capabilities of the underlying foundation models. Further discussion is provided in Appendix C.

\section*{Acknowledgments} The authors gratefully acknowledge the Kotak-IISc AI/ML Centre (KIAC) for the GPU resources that enabled this research.

{
    \small
    \bibliographystyle{ieeenat_fullname}
    \bibliography{ref}

\begin{thebibliography}{51}
\providecommand{\natexlab}[1]{#1}
\providecommand{\url}[1]{\texttt{#1}}
\expandafter\ifx\csname urlstyle\endcsname\relax
  \providecommand{\doi}[1]{doi: #1}\else
  \providecommand{\doi}{doi: \begingroup \urlstyle{rm}\Url}\fi

\bibitem[Achlioptas et~al.(2023)Achlioptas, Huang, Sung, Tulyakov, and
  Guibas]{achlioptas2023shapetalk}
Panos Achlioptas, Ian Huang, Minhyuk Sung, Sergey Tulyakov, and Leonidas
  Guibas.
\newblock {ShapeTalk}: A language dataset and framework for {3D} shape edits
  and deformations.
\newblock In \emph{Proceedings of the IEEE/CVF Conference on Computer Vision
  and Pattern Recognition}, pages 12685--12694, 2023.

\bibitem[Alrashedy et~al.(2025)Alrashedy, Tambwekar, Zaidi, Langwasser, Xu, and
  Gombolay]{al2025cadcodeverify}
Kamel Alrashedy, Pradyumna Tambwekar, Zulfiqar~Haider Zaidi, Megan Langwasser,
  Wei Xu, and Matthew Gombolay.
\newblock Generating {CAD} code with vision-language models for {3D} designs.
\newblock In \emph{International Conference on Learning Representations}, pages
  52236--52262, 2025.

\bibitem[{Anthropic}(2025)]{anthropic2025claudehaiku45}
{Anthropic}.
\newblock {Claude Haiku 4.5} system card.
\newblock Technical report, Anthropic, 2025.

\bibitem[{Anthropic}(2026)]{anthropic2026claudeSonnet46SystemCard}
{Anthropic}.
\newblock {Claude Sonnet 4.6} system card.
\newblock Technical report, Anthropic, 2026.

\bibitem[Behravan et~al.(2025)Behravan, Haghani, and
  Gra{\v{c}}anin]{behravan2025augmented}
Majid Behravan, Maryam Haghani, and Denis Gra{\v{c}}anin.
\newblock Transcending dimensions using generative {AI}: Real-time {3D} model
  generation in augmented reality.
\newblock In \emph{International Conference on Human-Computer Interaction},
  pages 13--32. Springer, 2025.

\bibitem[Besl and McKay(1992)]{besl1992method}
Paul~J Besl and Neil~D McKay.
\newblock A method for registration of {3-D} shapes.
\newblock \emph{IEEE Transactions on Pattern Analysis and Machine
  Intelligence}, 14\penalty0 (2):\penalty0 239--256, 1992.

\bibitem[Bradley and Terry(1952)]{bradley1952rank}
Ralph~Allan Bradley and Milton~E. Terry.
\newblock Rank analysis of incomplete block designs: {I}. {The} method of
  paired comparisons.
\newblock \emph{Biometrika}, 39\penalty0 (3/4):\penalty0 324--345, 1952.

\bibitem[Chang et~al.(2015)Chang, Funkhouser, Guibas, Hanrahan, Huang, Li,
  Savarese, Savva, Song, Su, et~al.]{chang2015shapenet}
Angel~X Chang, Thomas Funkhouser, Leonidas Guibas, Pat Hanrahan, Qixing Huang,
  Zimo Li, Silvio Savarese, Manolis Savva, Shuran Song, Hao Su, et~al.
\newblock {ShapeNet}: An information-rich {3D} model repository.
\newblock \emph{arXiv preprint arXiv:1512.03012}, 2015.

\bibitem[Chen et~al.(2025)Chen, Wei, Chen, Zhang, Yang, Zhang, Yang, Foo, Lin,
  Huang, and Liu]{chen2025cadcrafter}
Cheng Chen, Jiacheng Wei, Tianrun Chen, Chi Zhang, Xiaofeng Yang, Shangzhan
  Zhang, Bingchen Yang, Chuan-Sheng Foo, Guosheng Lin, Qixing Huang, and Fayao
  Liu.
\newblock {CADCrafter}: Generating computer-aided design models from
  unconstrained images.
\newblock In \emph{Proceedings of the IEEE/CVF Conference on Computer Vision
  and Pattern Recognition}, pages 11073--11082, 2025.

\bibitem[Chen et~al.(2024)Chen, Lan, Chen, Zhou, and Pan]{chen2024mvdrag3d}
Honghua Chen, Yushi Lan, Yongwei Chen, Yifan Zhou, and Xingang Pan.
\newblock {MvDrag3D}: Drag-based creative {3D} editing via multi-view
  generation-reconstruction priors.
\newblock \emph{arXiv preprint arXiv:2410.16272}, 2024.

\bibitem[Comanici et~al.(2025)]{gemini25}
Gheorghe Comanici et~al.
\newblock Gemini 2.5: Pushing the frontier with advanced reasoning,
  multimodality, long context, and next generation agentic capabilities.
\newblock \emph{arXiv preprint arXiv:2507.06261}, 2025.

\bibitem[{DeepSeek-AI}(2026)]{deepseek2026v4}
{DeepSeek-AI}.
\newblock {DeepSeek-V4}: Towards highly efficient million-token context
  intelligence.
\newblock \emph{arXiv preprint arXiv:2606.19348}, 2026.

\bibitem[Doris et~al.(2026)Doris, Alam, Nobari, and Ahmed]{doris2026cadcoder}
Anna~C. Doris, Md~Ferdous Alam, Amin~Heyrani Nobari, and Faez Ahmed.
\newblock {CAD-Coder}: An open-source vision-language model for computer-aided
  design code generation.
\newblock \emph{Journal of Mechanical Design}, 148\penalty0 (7):\penalty0
  071702, 2026.

\bibitem[Du et~al.(2024)Du, Chen, Zan, Li, Wang, Song, Li, Hu, and
  Wang]{du2024blenderllm}
Yuhao Du, Shunian Chen, Wenbo Zan, Peizhao Li, Mingxuan Wang, Dingjie Song, Bo
  Li, Yan Hu, and Benyou Wang.
\newblock {BlenderLLM}: Training large language models for computer-aided
  design with self-improvement.
\newblock \emph{arXiv preprint arXiv:2412.14203}, 2024.

\bibitem[Fang et~al.(2025)Fang, Shen, Wang, Tsai, Yang, Zhou, Ding, Igarashi,
  and Yang]{fang2025meshllm}
Shuangkang Fang, I-Chao Shen, Yufeng Wang, Yi-Hsuan Tsai, Yi Yang, Shuchang
  Zhou, Wenrui Ding, Takeo Igarashi, and Ming-Hsuan Yang.
\newblock {MeshLLM}: Empowering large language models to progressively
  understand and generate {3D} mesh.
\newblock In \emph{Proceedings of the IEEE/CVF International Conference on
  Computer Vision}, pages 14061--14072, 2025.

\bibitem[Fukaya et~al.(2025)Fukaya, Daylamani-Zad, and
  Agius]{fukaya2025intelligent}
Kaisei Fukaya, Damon Daylamani-Zad, and Harry Agius.
\newblock Intelligent generation of graphical game assets: A conceptual
  framework and systematic review of the state of the art.
\newblock \emph{ACM Computing Surveys}, 57\penalty0 (5):\penalty0 1--38, 2025.

\bibitem[Jun and Nichol(2023)]{jun2023shap-e}
Heewoo Jun and Alex Nichol.
\newblock {Shap-E}: Generating conditional {3D} implicit functions.
\newblock \emph{arXiv preprint arXiv:2305.02463}, 2023.

\bibitem[Li et~al.(2025)Li, Ma, Li, Lou, Zhou, and Zhou]{li2025cadllama}
Jiahao Li, Weijian Ma, Xueyang Li, Yunzhong Lou, Guichun Zhou, and Xiangdong
  Zhou.
\newblock {CAD-Llama}: Leveraging large language models for computer-aided
  design parametric {3D} model generation.
\newblock In \emph{Proceedings of the IEEE/CVF Conference on Computer Vision
  and Pattern Recognition}, pages 18563--18573, 2025.

\bibitem[Meng et~al.(2022)Meng, He, Song, Song, Wu, Zhu, and
  Ermon]{meng2022sdedit}
Chenlin Meng, Yutong He, Yang Song, Jiaming Song, Jiajun Wu, Jun-Yan Zhu, and
  Stefano Ermon.
\newblock {SDEdit}: Guided image synthesis and editing with stochastic
  differential equations.
\newblock In \emph{International Conference on Learning Representations}, 2022.

\bibitem[{Moonshot AI}(2026)]{moonshotai2026kimik26}
{Moonshot AI}.
\newblock {Kimi K2.6}: Advancing open-source coding.
\newblock \url{https://www.kimi.com/blog/kimi-k2-6.html}, 2026.

\bibitem[Mou et~al.(2024)Mou, Wang, Song, Shan, and Zhang]{mou2024diffeditor}
Chong Mou, Xintao Wang, Jiechong Song, Ying Shan, and Jian Zhang.
\newblock {DiffEditor}: Boosting accuracy and flexibility on diffusion-based
  image editing.
\newblock In \emph{Proceedings of the IEEE/CVF Conference on Computer Vision
  and Pattern Recognition}, pages 8488--8497, 2024.

\bibitem[Mueller et~al.(2025)Mueller, Uenlue, Schmidt, Kollovieh, Fan,
  G{\"u}nnemann, and Mikelsons]{mueller2025geodiffusion}
Phillip Mueller, Talip Uenlue, Sebastian Schmidt, Marcel Kollovieh, Jiajie Fan,
  Stephan G{\"u}nnemann, and Lars Mikelsons.
\newblock {GeoDiffusion}: A training-free framework for accurate {3D} geometric
  conditioning in image generation.
\newblock In \emph{Proceedings of the IEEE/CVF International Conference on
  Computer Vision}, pages 6374--6384, 2025.

\bibitem[Nichol et~al.(2022)Nichol, Jun, Dhariwal, Mishkin, and
  Chen]{nichol2022pointe}
Alex Nichol, Heewoo Jun, Prafulla Dhariwal, Pamela Mishkin, and Mark Chen.
\newblock {Point-E}: A system for generating {3D} point clouds from complex
  prompts.
\newblock \emph{arXiv preprint arXiv:2212.08751}, 2022.

\bibitem[{OpenAI}(2024)]{openai2024gpt4omini}
{OpenAI}.
\newblock {GPT-4o mini}: Advancing cost-efficient intelligence.
\newblock
  \url{https://openai.com/index/gpt-4o-mini-advancing-cost-efficient-intelligence/},
  2024.

\bibitem[{OpenAI}(2025)]{openai2025gpt5}
{OpenAI}.
\newblock Introducing {GPT-5}.
\newblock \url{https://openai.com/index/introducing-gpt-5/}, 2025.

\bibitem[{OpenAI}(2026)]{openai2026gpt54}
{OpenAI}.
\newblock {GPT-5.4 Thinking} system card.
\newblock Technical report, OpenAI, 2026.
\newblock Section~6: {GPT-5.4 mini}.

\bibitem[Piegl and Tiller(2012)]{piegl2012nurbsbook}
Les Piegl and Wayne Tiller.
\newblock \emph{The {NURBS} {Book}}.
\newblock Springer Science \& Business Media, 2012.

\bibitem[{Qwen Team}(2026)]{qwen2026qwen35b}
{Qwen Team}.
\newblock {Qwen3.6-35B-A3B}: Agentic coding power, now open to all.
\newblock \url{https://qwen.ai/blog?id=qwen3.6-35b-a3b}, 2026.

\bibitem[Radford et~al.(2021)Radford, Kim, Hallacy, Ramesh, Goh, Agarwal,
  Sastry, Askell, Mishkin, Clark, Krueger, and Sutskever]{radford2021clip}
Alec Radford, Jong~Wook Kim, Chris Hallacy, Aditya Ramesh, Gabriel Goh,
  Sandhini Agarwal, Girish Sastry, Amanda Askell, Pamela Mishkin, Jack Clark,
  Gretchen Krueger, and Ilya Sutskever.
\newblock Learning transferable visual models from natural language
  supervision.
\newblock In \emph{Proceedings of the 38th International Conference on Machine
  Learning}, pages 8748--8763. PMLR, 2021.

\bibitem[Sella et~al.(2026)Sella, Phung, Amiel, Litany, Patashnik, and
  Averbuch-Elor]{sella2026proxe}
Etai Sella, Hao Phung, Nitay Amiel, Or Litany, Or Patashnik, and Hadar
  Averbuch-Elor.
\newblock {Prox-E}: Fine-grained {3D} shape editing via primitive-based
  abstractions.
\newblock In \emph{SIGGRAPH Conference Papers '26}, pages 1--12. ACM, 2026.

\bibitem[Shi et~al.(2024)Shi, Wang, Ye, Mai, Li, and Yang]{shi2024mvdream}
Yichun Shi, Peng Wang, Jianglong Ye, Long Mai, Kejie Li, and Xiao Yang.
\newblock {MVDream}: Multi-view diffusion for {3D} generation.
\newblock In \emph{International Conference on Learning Representations}, 2024.

\bibitem[Shu et~al.(2019)Shu, Park, and Kwon]{shu2019fpd}
Dong~Wook Shu, Sung~Woo Park, and Junseok Kwon.
\newblock {3D} point cloud generative adversarial network based on tree
  structured graph convolutions.
\newblock In \emph{Proceedings of the IEEE/CVF International Conference on
  Computer Vision}, pages 3859--3868, 2019.

\bibitem[Sinha et~al.(2025)Sinha, Khan, Usama, Sam, Stricker, Ali, and
  Afzal]{sinha2025marvel}
Sankalp Sinha, Mohammad~Sadil Khan, Muhammad Usama, Shino Sam, Didier Stricker,
  Sk~Aziz Ali, and Muhammad~Zeshan Afzal.
\newblock {MARVEL-40M+}: Multi-level visual elaboration for high-fidelity
  text-to-{3D} content creation.
\newblock In \emph{Proceedings of the IEEE/CVF Conference on Computer Vision
  and Pattern Recognition}, pages 8105--8116, 2025.

\bibitem[Stojanov et~al.(2021)Stojanov, Thai, and Rehg]{stojanov2021using}
Stefan Stojanov, Anh Thai, and James~M Rehg.
\newblock Using shape to categorize: Low-shot learning with an explicit shape
  bias.
\newblock In \emph{Proceedings of the IEEE/CVF Conference on Computer Vision
  and Pattern Recognition}, pages 1798--1808, 2021.

\bibitem[Usama et~al.(2026)Usama, Khan, Stricker, and Afzal]{usama2026nurbgen}
Muhammad Usama, Mohammad~Sadil Khan, Didier Stricker, and Muhammad~Zeshan
  Afzal.
\newblock {NURBGen}: High-fidelity text-to-{CAD} generation through
  {LLM}-driven {NURBS} modeling.
\newblock In \emph{Proceedings of the AAAI Conference on Artificial
  Intelligence}, pages 9603--9611, 2026.

\bibitem[Vinker et~al.(2025)Vinker, Shaham, Zheng, Zhao, Fan, and
  Torralba]{vinker2025sketchagent}
Yael Vinker, Tamar~Rott Shaham, Kristine Zheng, Alex Zhao, Judith~E Fan, and
  Antonio Torralba.
\newblock {SketchAgent}: Language-driven sequential sketch generation.
\newblock In \emph{Proceedings of the IEEE/CVF Conference on Computer Vision
  and Pattern Recognition}, pages 23355--23368, 2025.

\bibitem[Wang et~al.(2024)Wang, Lorraine, Wang, Su, Zhu, Fidler, and
  Zeng]{wang2024llama-mesh}
Zhengyi Wang, Jonathan Lorraine, Yikai Wang, Hang Su, Jun Zhu, Sanja Fidler,
  and Xiaohui Zeng.
\newblock {LLaMA-Mesh}: Unifying {3D} mesh generation with language models.
\newblock \emph{arXiv preprint arXiv:2411.09595}, 2024.

\bibitem[Worchel and Alexa(2023)]{worchel2023differentiable}
Markus Worchel and Marc Alexa.
\newblock Differentiable rendering of parametric geometry.
\newblock \emph{ACM Transactions on Graphics (TOG)}, 42\penalty0 (6):\penalty0
  1--18, 2023.

\bibitem[Wu et~al.(2021)Wu, Xiao, and Zheng]{wu2021deepcad}
Rundi Wu, Chang Xiao, and Changxi Zheng.
\newblock {DeepCAD}: A deep generative network for computer-aided design
  models.
\newblock In \emph{Proceedings of the IEEE/CVF International Conference on
  Computer Vision}, pages 6772--6782, 2021.

\bibitem[Wu et~al.(2023)Wu, Zhang, Fu, Wang, Ren, Pan, Wu, Yang, Wang, Qian,
  Lin, and Liu]{wu2023omniobject3d}
Tong Wu, Jiarui Zhang, Xiao Fu, Yuxin Wang, Jiawei Ren, Liang Pan, Wayne Wu,
  Lei Yang, Jiaqi Wang, Chen Qian, Dahua Lin, and Ziwei Liu.
\newblock {OmniObject3D}: Large-vocabulary {3D} object dataset for realistic
  perception, reconstruction and generation.
\newblock In \emph{Proceedings of the IEEE/CVF Conference on Computer Vision
  and Pattern Recognition}, pages 803--814, 2023.

\bibitem[Xiang et~al.(2025)Xiang, Lv, Xu, Deng, Wang, Zhang, Chen, Tong, and
  Yang]{xiang2025trellis}
Jianfeng Xiang, Zelong Lv, Sicheng Xu, Yu Deng, Ruicheng Wang, Bowen Zhang,
  Dong Chen, Xin Tong, and Jiaolong Yang.
\newblock Structured {3D} latents for scalable and versatile {3D} generation.
\newblock In \emph{Proceedings of the IEEE/CVF Conference on Computer Vision
  and Pattern Recognition}, pages 21469--21480, 2025.

\bibitem[Xiang et~al.(2026)Xiang, Chen, Xu, Wang, Lv, Deng, Zhu, Dong, Zhao,
  Yuan, et~al.]{xiang2026native}
Jianfeng Xiang, Xiaoxue Chen, Sicheng Xu, Ruicheng Wang, Zelong Lv, Yu Deng,
  Hongyuan Zhu, Yue Dong, Hao Zhao, Nicholas~Jing Yuan, et~al.
\newblock Native and compact structured latents for {3D} generation.
\newblock In \emph{Proceedings of the IEEE/CVF Conference on Computer Vision
  and Pattern Recognition}, pages 14419--14429, 2026.

\bibitem[{Xiaomi MiMo Team}(2026)]{xiaomi2026mimov25}
{Xiaomi MiMo Team}.
\newblock {MiMo-V2.5}.
\newblock \url{https://huggingface.co/collections/XiaomiMiMo/mimo-v25}, 2026.

\bibitem[Xu et~al.(2024)Xu, Lambourne, Jayaraman, Wang, Willis, and
  Furukawa]{xu2024brepgen}
Xiang Xu, Joseph Lambourne, Pradeep Jayaraman, Zhengqing Wang, Karl Willis, and
  Yasutaka Furukawa.
\newblock {BrepGen}: A {B-rep} generative diffusion model with structured
  latent geometry.
\newblock \emph{ACM Transactions on Graphics (TOG)}, 43\penalty0 (4):\penalty0
  1--14, 2024.

\bibitem[Xue et~al.(2023)Xue, Gao, Xing, Mart{\'\i}n-Mart{\'\i}n, Wu, Xiong,
  Xu, Niebles, and Savarese]{xue2023ulip}
Le Xue, Mingfei Gao, Chen Xing, Roberto Mart{\'\i}n-Mart{\'\i}n, Jiajun Wu,
  Caiming Xiong, Ran Xu, Juan~Carlos Niebles, and Silvio Savarese.
\newblock {ULIP}: Learning a unified representation of language, images, and
  point clouds for {3D} understanding.
\newblock In \emph{Proceedings of the IEEE/CVF Conference on Computer Vision
  and Pattern Recognition}, pages 1179--1189, 2023.

\bibitem[Yamada et~al.(2025)Yamada, Chandu, Lin, Hessel, Yildirim, and
  Choi]{yamada2025l3go}
Yutaro Yamada, Khyathi Chandu, Bill~Yuchen Lin, Jack Hessel, Ilker Yildirim,
  and Yejin Choi.
\newblock {L3GO}: Language agents with chain-of-{3D}-thoughts for generating
  unconventional objects.
\newblock In \emph{Proceedings of the 2025 Conference of the Nations of the
  {A}mericas Chapter of the {A}ssociation for {C}omputational {L}inguistics:
  Human Language Technologies (System Demonstrations)}, pages 456--469, 2025.

\bibitem[Ye et~al.(2025)Ye, Wang, Zhao, Xie, and Zhu]{wang2026shapellm}
Junliang Ye, Zhengyi Wang, Ruowen Zhao, Shenghao Xie, and Jun Zhu.
\newblock {ShapeLLM-Omni}: A native multimodal {LLM} for {3D} generation and
  understanding.
\newblock \emph{Advances in Neural Information Processing Systems},
  38:\penalty0 89171--89204, 2025.

\bibitem[You et~al.(2026)You, Zhao, Zhang, Xu, and Fua]{you2026physgen}
Yingxuan You, Chen Zhao, Hantao Zhang, Ming Xu, and Pascal Fua.
\newblock {PhysGen}: Physically grounded {3D} shape generation for industrial
  design.
\newblock In \emph{Proceedings of the IEEE/CVF Conference on Computer Vision
  and Pattern Recognition}, pages 27209--27218, 2026.

\bibitem[Zhang et~al.(2018)Zhang, Isola, Efros, Shechtman, and
  Wang]{zhang2018lpips}
Richard Zhang, Phillip Isola, Alexei~A. Efros, Eli Shechtman, and Oliver Wang.
\newblock The unreasonable effectiveness of deep features as a perceptual
  metric.
\newblock In \emph{Proceedings of the IEEE Conference on Computer Vision and
  Pattern Recognition}, pages 586--595, 2018.

\bibitem[Zhang et~al.(2025)Zhang, Jiang, Li, and Deng]{zhang2026shapecraft}
Shuyuan Zhang, ChenHan Jiang, Zuoou Li, and Jiankang Deng.
\newblock {ShapeCraft}: {LLM} agents for structured, textured and interactive
  {3D} modeling.
\newblock \emph{Advances in Neural Information Processing Systems},
  38:\penalty0 65116--65144, 2025.

\bibitem[Zhou et~al.(2026)Zhou, Li, Lyu, Song, Huang, Zheng, Rupprecht,
  Vedaldi, and Wu]{zhou2026articraft}
Matt Zhou, Ruining Li, Xiaoyang Lyu, Zhaomou Song, Zhening Huang, Chuanxia
  Zheng, Christian Rupprecht, Andrea Vedaldi, and Shangzhe Wu.
\newblock Articraft: An agentic system for scalable articulated {3D} asset
  generation.
\newblock \emph{arXiv preprint arXiv:2605.15187}, 2026.

\end{thebibliography}
}

\clearpage
\appendix
\twocolumn[{%
  \null\vskip .375in
  \begin{center}
    {\Large \bf Supplementary Materials for \textit{SurfScaff3D}: Bridging Geometric and Semantic Representations for LLM-based 3D Object Generation \par}
  \end{center}
  \vspace*{24pt}
}]

The following is the appendix for the paper ``SurfScaff3D: Bridging Geometric and Semantic
Representations for LLM-based 3D Object Generation''. It is organized as follows. Section A provides the technical details for the implementation and evaluation of SurfScaff3D. In Section B, we report additional qualitative results not included in the main paper. Section C discusses the limitations of SurfScaff3D. Section D discusses comparison with related works and Section E evaluates it with other backbones.

\section{Technical Setup}
\subsection{Implementation Details}

\paragraph{Generator Agent}
For SurfScaff3D, we use Claude Sonnet 4.6~\cite{anthropic2026claudeSonnet46SystemCard} as the generator, configured with a maximum of 8,000 extended-thinking tokens and a sampling temperature of 1.0 (default). For SurfScaff3D-K, we use Kimi K2.6~\cite{moonshotai2026kimik26} with medium reasoning effort and temperature 0.0. For SurfScaff3D-Q, we use Qwen3.6-35B-A3B~\cite{qwen2026qwen35b} with temperature 0.0 and the model's default reasoning behaviour.

The code execution tool runs Python in an isolated sandbox or a subprocess and returns stdout and stderr, enabling the agent to compute coordinates, verify connectivity, and write the final HPS representation to \texttt{final\_output.txt}. The rendering tool renders the current patches and returns a six-view image montage (four three-quarter views, one top-down, and one bottom-up), allowing the agent to visually inspect geometry before finalization. Eight in-context library examples are prepended to the user prompt and shared across refinement turns and across all objects generated in the same batch. Each generator turn allows up to 20 tool-call steps. Transient API failures are retried up to three times using exponential backoff capped at 20 seconds.

\begin{figure}[h]
    \centering
    \includegraphics[width=\linewidth]{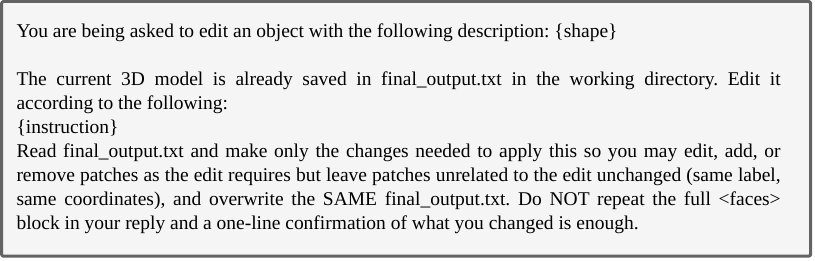}
    \caption{Prompt for editing existing HPS representations.}
    \label{fig:edit_xtra}
\end{figure}

\begin{figure*}
    \centering
    \includegraphics[width=\linewidth]{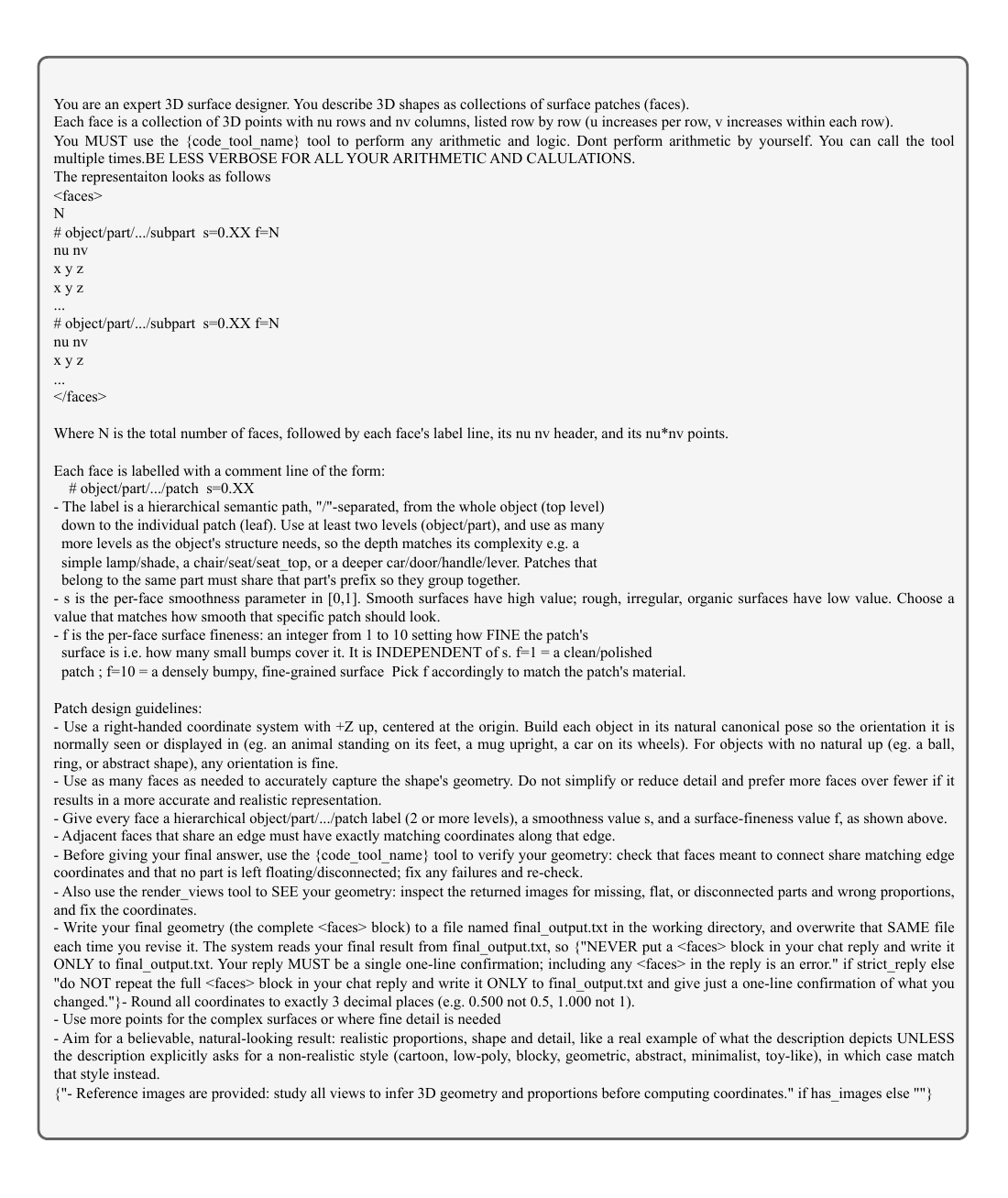}
    \caption{System Prompt for SurfScaff3D Generator Agent.}
    \label{fig:gen_sys_prompt}
\end{figure*}

\begin{figure*}
    \centering
    \includegraphics[width=0.85\linewidth]{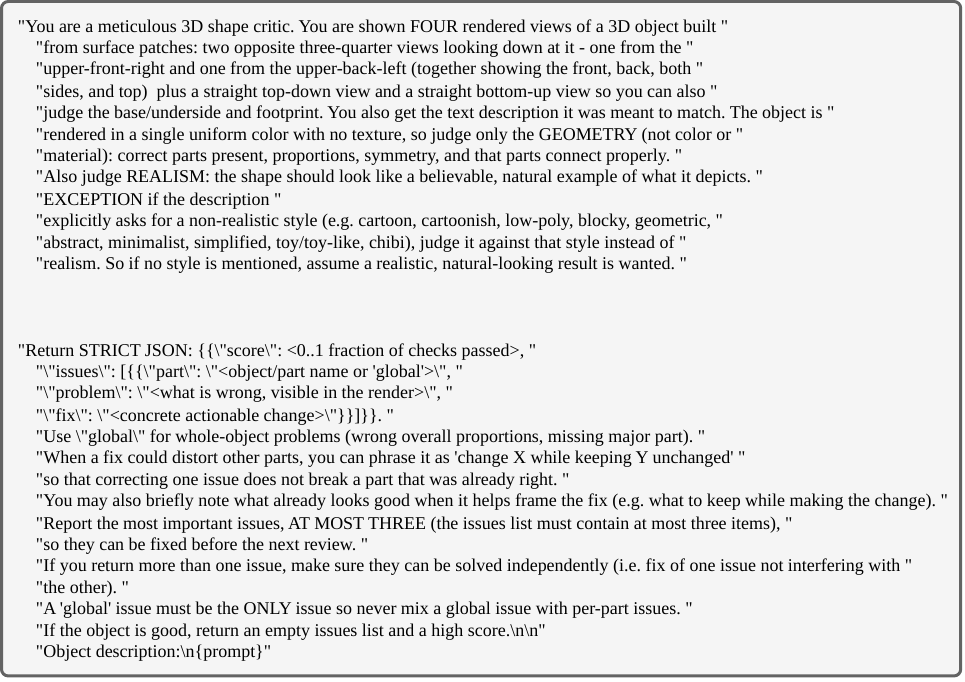}
    \caption{System prompt for SurfScaff3D critic agent.}
    \label{fig:critic_sys_prompt}
\end{figure*}

When reference images are provided, we append the instruction ``Reference images are provided: study all views to infer 3D geometry and proportions before computing coordinates.'' to the generator's system prompt, and the user prompt introduces the attached images. For editing existing HPS representations, the edit instruction is inserted into the prompt shown in \autoref{fig:edit_xtra}. Finally, to facilitate better reproducibility, we have provided the generator system prompt in \autoref{fig:gen_sys_prompt} and the system prompt for the critic agent in \autoref{fig:critic_sys_prompt}.

\begin{figure*}
    \centering
\includegraphics[width=0.85\linewidth]{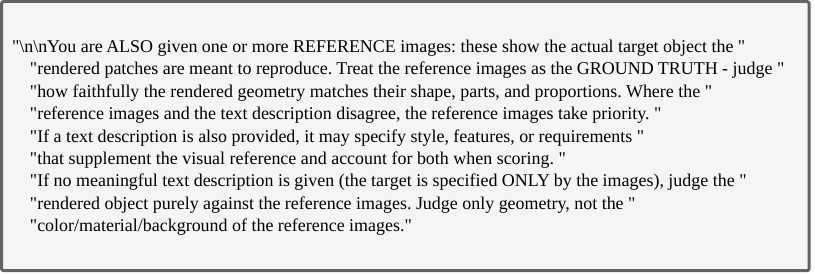}
    \caption{Prompt appended to SurfScaff3D critic agent when an image is provided.}
    \label{fig:image_critic_sys_prompt}
\end{figure*}

\paragraph{Critic Agent}
We use Gemini 2.5 Pro~\cite{gemini25} as the critic for all three variants, with temperature 0.0 and medium thinking. The critic receives four rendered views of the current HPS geometry (two three-quarter views, one top-down, and one bottom-up), rendered in a single uniform colour to evaluate geometry only, together with the original text prompt. For image-conditioned generation, the reference images are additionally provided, and the prompt in \autoref{fig:image_critic_sys_prompt} is appended to the critic's system prompt.

The critic returns a structured JSON response containing a scalar score in $[0,1]$ and up to three issues, each consisting of a target part, the observed problem, and a concrete fix. Invalid or empty responses are retried up to four times with exponential backoff.

The generator and critic alternate for up to five iterations. If the critic score exceeds 0.9, the loop terminates early; otherwise, identified issues are appended to the generator context as structured feedback. The system tracks the highest-scoring geometry across iterations and restores it as the final output.

\paragraph{HPS Visualization}
We convert HPS patches into B-spline surfaces following \textit{The NURBS Book}~\cite{piegl2012nurbsbook}, remapping per-patch $s_f$ to $[0.05, 1.0]$, with $s_f \geq 0.9$ clamped to $1.0$, $s_f \in [0.6, 0.9)$ stretched linearly onto $[0.6, 1.0)$, and $s_f < 0.6$ kept unchanged. We clamp $\omega_f$ to $[1, 10]$. Surfaces are then tessellated into meshes at a resolution of $60 \times 60$ samples per face, providing sufficient detail to capture surface variations while remaining efficient for complex meshes with many faces. This configuration is used for all qualitative and quantitative results in the paper.

\subsection{Evaluation Setup} 

\paragraph{Generation}
To evaluate text-to-3D generation, we use prompts from MARVEL-40M+~\cite{sinha2025marvel} for objects from ShapeNet-Core~\cite{chang2015shapenet}, OmniObject3D~\cite{wu2023omniobject3d}, and Toys4K~\cite{stojanov2021using}. MARVEL-40M+ provides prompts ranging from highly concise (level-5) to highly detailed (level-1); we use level-3 prompts as they provide a practical balance between detail and conciseness. Texture-related information is removed from the prompts using Claude Haiku 4.5~\cite{anthropic2025claudehaiku45}.

For evaluation, we sample one object from each class in ShapeNetCore and Toys4K, and one object from each of a subset of classes in OmniObject3D, to keep evaluation and ablation costs down. This results in a total of 55 objects for ShapeNetCore, 105 objects for Toys4K and 71 for OmniObject3D, providing a diverse evaluation set spanning CAD-like, structural, and organic objects.

\paragraph{Quantitative Metrics}
For CLIP evaluation, we render 10 views of each 3D asset and pass them through the CLIP image encoder. For ULIP evaluation, we sample a 10,000-point point cloud from the generated mesh and encode it using ULIP’s point cloud encoder. In both cases, we compute the average cosine similarity between the image (or point-cloud) embeddings and the text embeddings of the corresponding MARVEL-40M+ level-4 prompts. Although generation uses level-3 prompts, we use level-4 prompts for evaluation because CLIP is limited to a 77-token context window, while level-4 prompts retain most information without truncation.

For VQA evaluation, we use GPT-5 mini~\cite{openai2025gpt5} to generate five yes/no questions from the input prompt, targeting key semantic attributes of the object. The same question set is used across all evaluated models. We provide five rendered views of the generated asset and ask GPT-5 mini to answer the previously generated questions. The final VQA score is reported as the average accuracy. The failure rate is the percentage of objects across the three datasets for which a method fails to produce a renderable mesh.

\paragraph{Image-to-3D Evaluation}
Each generated mesh is registered to its ground-truth mesh via a 24-rotation search followed by the Iterative Closest Point (ICP) algorithm~\cite{besl1992method}. We then report geometric accuracy against ground truth (Chamfer distance, F-score@0.05), how closely the output matches the conditioning image (CLIP-I), and how accurately the model reconstructs object parts not shown in the input image, measured by rendering the generated mesh from the viewpoints not seen in the input and comparing these renders to the ground-truth object's renders from the same viewpoints (LPIPS).

\paragraph{Elo Evaluation}
For Elo evaluation, we render two image sets per object: i) a four-view grid containing front-right, back-left, top-down, and bottom-up views arranged in a $2\times2$ layout, and ii) a close-up ($2\times$ zoom) surface-normal rendering of the same views. We use GPT-5.4-mini~\cite{openai2026gpt54} as the pairwise judge with low reasoning effort and a maximum of 4,000 output tokens. Each evaluation call receives four images (grid A, close-up A, grid B, and close-up B) together with the corresponding MARVEL-40M+ level-3 prompt. The judge is instructed to use the prompt only to identify the object category and assess whether the geometry and surface variations appear natural, while ignoring prompt alignment. For example, metallic surfaces are expected to be smooth and are penalized for irregular noise, whereas organic or wooden surfaces are credited for natural irregularity unless the prompt specifies otherwise.

The judge rates each method on three axes, i.e. structural coherence, detail quality, and surface appropriateness (integers from 1 to 10) and returns a preferred method, emitted as structured JSON under a strict schema. Each method pair is compared on every object for which at least one of the two methods produced a mesh, rather than only on objects for which all methods produced one, so that a single low-coverage baseline does not reduce the data available to the remaining pairs. Generation failures are counted as losses, i.e. if one method produced no mesh for a given object, the pair is scored as a win for the method that did, without invoking the judge; this prevents a method from improving its rating by declining to generate difficult objects. We fit a Bradley--Terry model~\cite{bradley1952rank} over all pairwise judgments, treating ties as half a win for each method. To keep the estimates finite when a method wins or loses every comparison against another, we add one virtual tie of weight 0.5 to every method pair. Strengths are then estimated by maximum likelihood and normalized to unit geometric mean. The resulting strengths are log-transformed and linearly rescaled such that a 400-point Elo difference corresponds to 10:1 odds of winning, with the mean rating in each dataset anchored at 1000.

\paragraph{Editing} We use the ShapeTalk~\cite{achlioptas2023shapetalk} dataset to evaluate 3D editing. We randomly sample one source object from each of the 30 classes and select three edit utterances per object, resulting in 90 source-object edit pairs.

Since ShapeTalk does not provide source prompts and our method requires an HPS representation for editing, we use Claude Haiku 4.5~\cite{anthropic2025claudehaiku45} to generate detailed prompts for reconstructing the source objects before editing. Each edit uses a single critic refinement iteration.

For CLIP and ULIP evaluation, we generate edited-object descriptions using Claude Haiku 4.5 with emphasis on the modified regions, following the same evaluation setup as generation. For VQA, following Prox-E~\cite{sella2026proxe}, we provide the original and edited renders together with the edit instruction and ask GPT-4o mini~\cite{openai2024gpt4omini} to determine whether the edit was correctly applied.

\section{Additional Qualitative Results}

\begin{figure*}
    \centering
    \includegraphics[width=\linewidth]{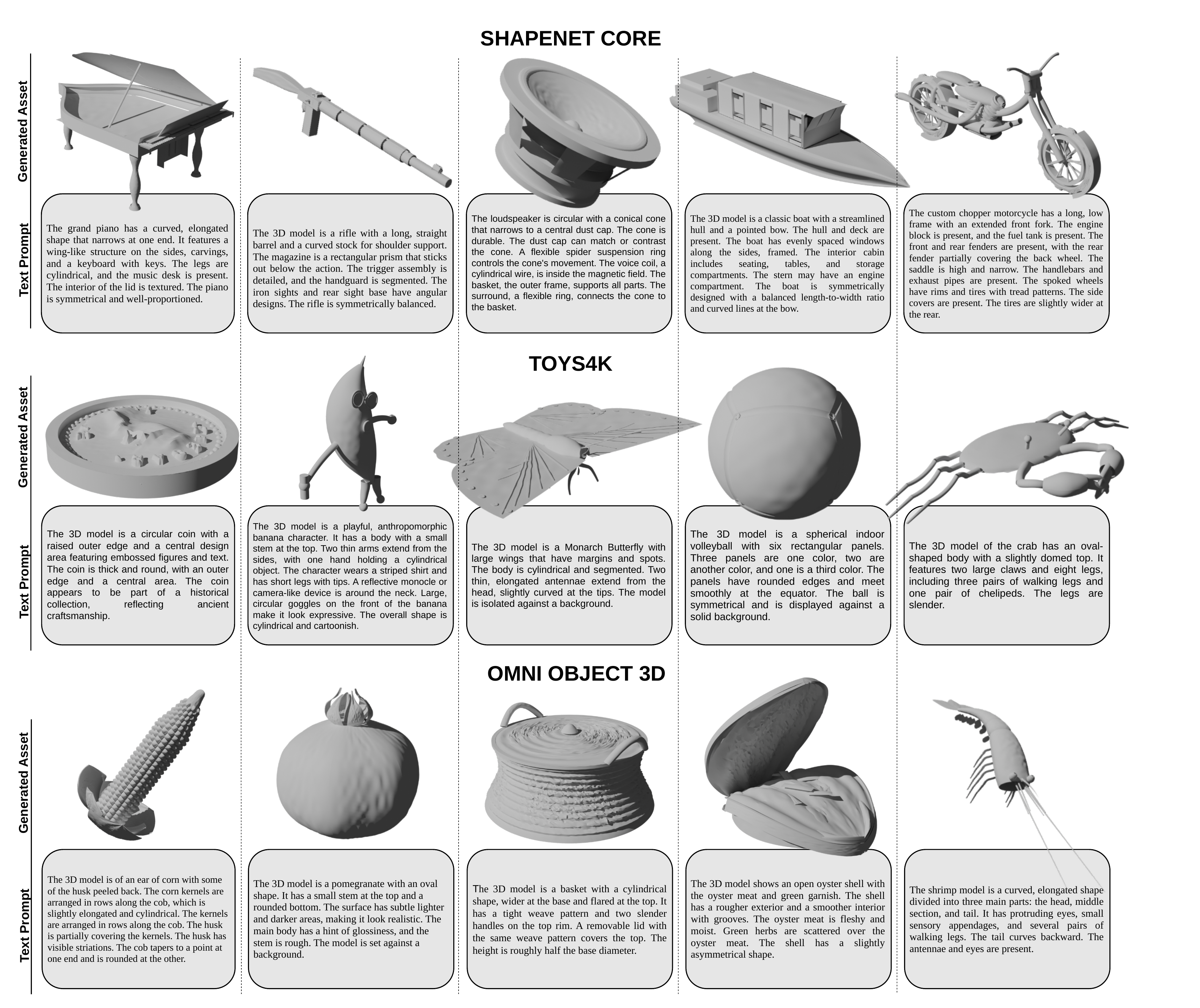}
    \caption{\textbf{Qualitative Samples for text-to-3D generation.} We show samples generated by SurfScaff3D from the object categories of ShapeNetCore \cite{chang2015shapenet}, Toys4K \cite{stojanov2021using} and OmniObject3D \cite{wu2023omniobject3d} datasets. The corresponding MARVEL-40M+ level-3 text prompts are displayed below each result.}
    \label{fig:text-to-3d}
\end{figure*}

\begin{figure*}
    \centering
    \includegraphics[width=\linewidth]{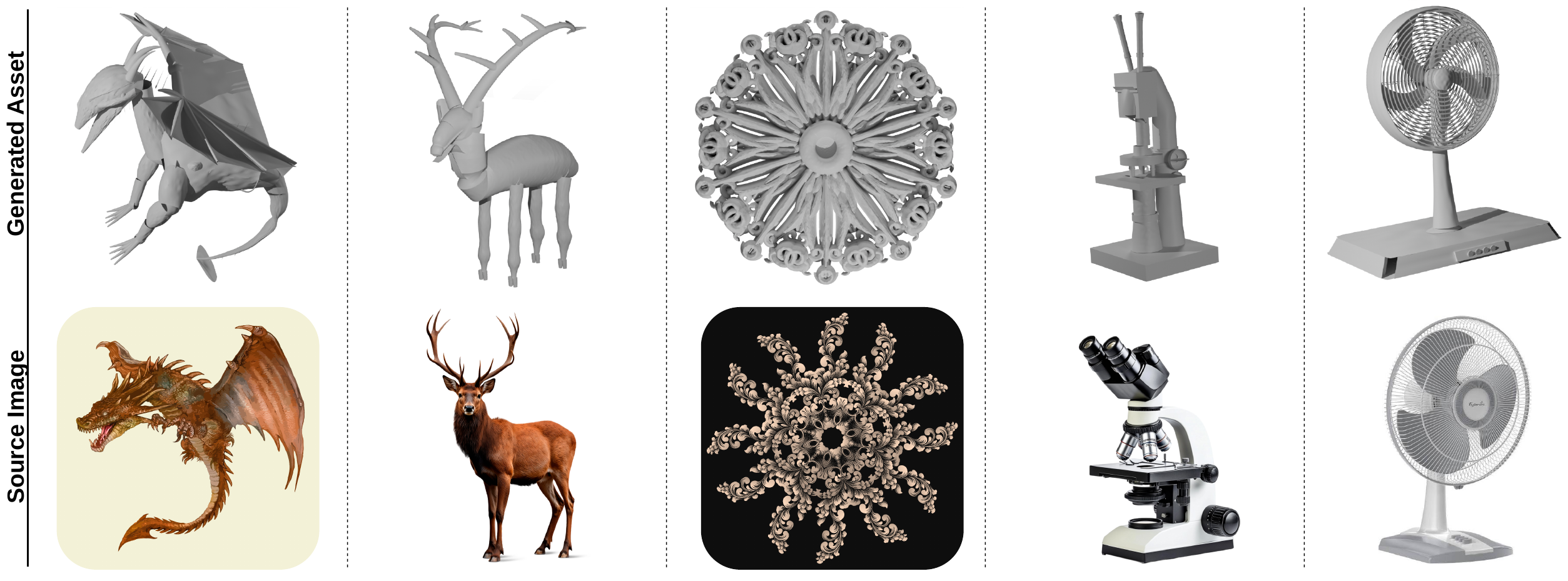}
    \caption{\textbf{Qualitative samples for image-to-3D.} We demonstrate the performance of SurfScaff3D on diverse image inputs. The source conditioning images are provided below each resulting 3D asset.}
    \label{fig:image-to-3d}
\end{figure*}

\begin{figure*}
    \centering
    \includegraphics[width=\linewidth]{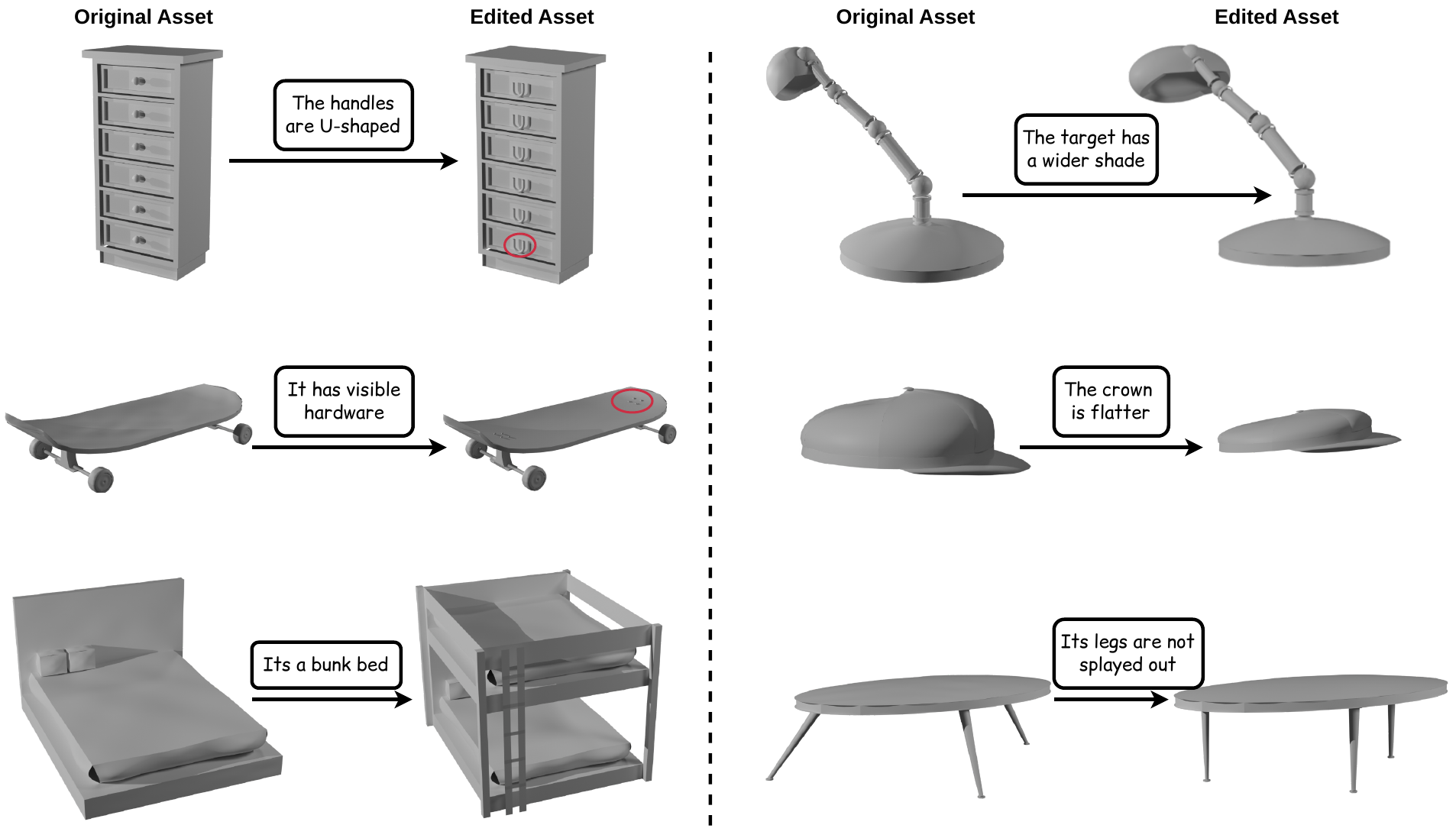}
    \caption{\textbf{Editing capabilities of our model.} We display the original 3D object alongside the edited result. The corresponding ShapeTalk-based text prompts used to guide the edits are shown on the arrows.}
    \label{fig:editing}
\end{figure*}

We present additional qualitative results omitted from the main paper due to space constraints. We showcase text-to-3D generation performance across the ShapeNetCore, OmniObject3D, and Toys4K datasets in Figure \ref{fig:text-to-3d}. Additionally, we provide examples of image-to-3D generation in Figure \ref{fig:image-to-3d} and demonstrate the editing capabilities of SurfScaff3D-K in Figure \ref{fig:editing}.

\section{Limitations}

\begin{figure*}
    \centering
    \includegraphics[width=\linewidth]{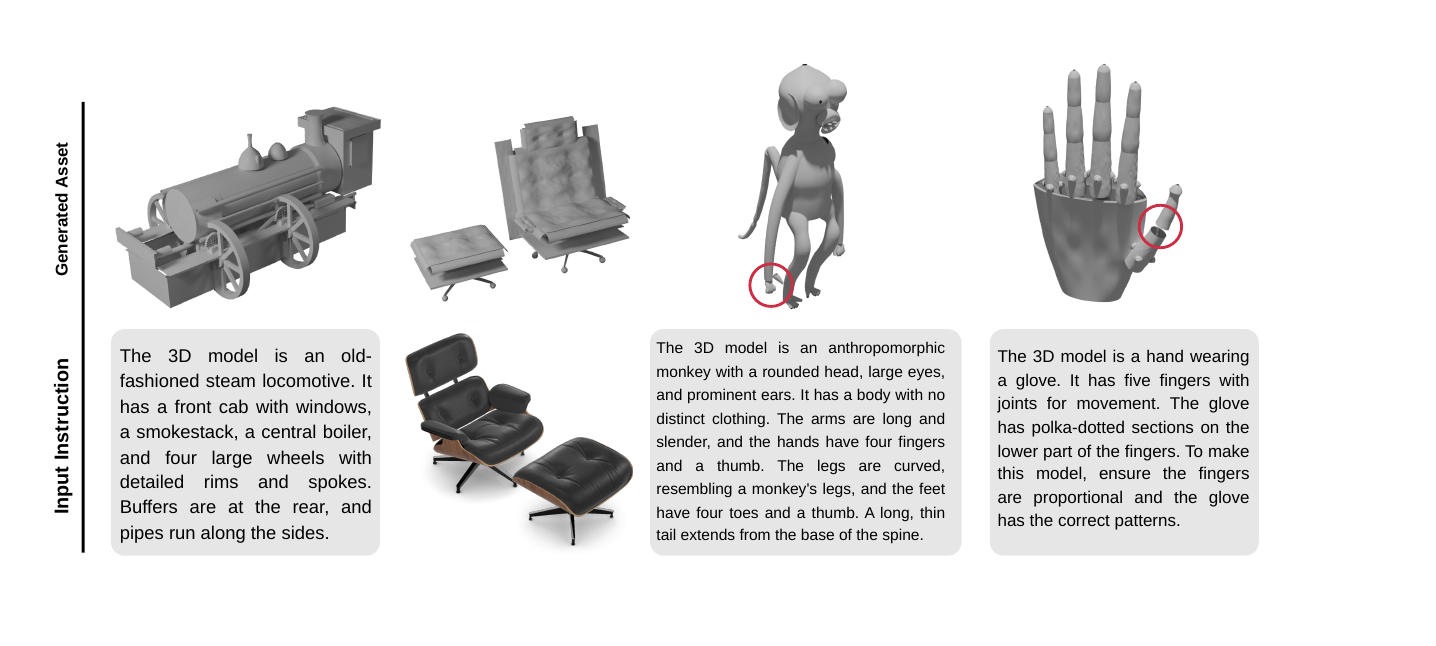}
    \caption{\textbf{Failure cases of SurfScaff3D.} The model can misjudge the orientation and relative positioning of object parts (left), and in some cases leaves parts detached from the body, producing visible gaps in the generated surface (right).}
    \label{fig:limitations}
\end{figure*}

SurfScaff3D relies on LLMs as its generative backbone and consequently inherits some of their limitations. As shown in \autoref{fig:limitations}, our model occasionally struggles with the precise spatial orientation and relative positioning of object components, which can result in logical misalignments in some 3D structures. Although the agent is explicitly instructed, before giving its final output, to verify that connecting faces share matching edge coordinates and that no part is left floating or disconnected, this verification is performed by the model itself and carries no formal guarantee. Consequently, in some cases parts may remain detached from the body, leaving gaps that make the generated surface visibly non-watertight, as shown in \autoref{fig:limitations}. Such failures can be addressed through human-in-the-loop refinement, as the explicit part-level structure of HPS enables a user to reconnect or reposition the affected patches without regenerating the object.

\section{Positioning Relative to Existing Frameworks}

\autoref{tab:framework_comparison} compares several generation and editing frameworks in terms of training requirements, structure, refinement, and editing capability. We observe that iterative refinement appears in five of the eleven prior frameworks, but is less often combined with an explicitly labelled part decomposition, so refinement tends to regenerate the program or re-optimise the shape globally rather than revise a named region.

Also, procedural and CAD-based frameworks expose a predefined vocabulary of primitives or modeling operations, so they can directly build only what that vocabulary and the underlying API support.

The editing-only methods also differ from ours in mechanism. Prox-E \cite{sella2026proxe} preserves unchanged regions by injecting the original shape's inverted latents, which the diffusion model then refines over the remaining denoising steps, while MvDrag3D~\cite{chen2024mvdrag3d} re-optimises the Gaussians through per-instance deformation networks followed by score-distillation-based appearance refinement. In both cases, unchanged regions are encouraged to remain consistent with the input rather than being guaranteed to remain numerically identical.

HPS addresses these together within a single representation. Its patches are explicitly labelled, so refinement can target a named region rather than regenerate the object. They are continuous spline surfaces rather than instances of a fixed operator set, allowing parts to take arbitrary continuous surface geometries rather than being restricted to predefined primitive types. Also, the editing acts directly on patch coordinates, so unedited parts can be left untouched rather than re-synthesised.

\begin{table*}[t]
\centering
\small
\setlength{\tabcolsep}{5pt}
\begin{tabular}{l ccccccc l}
\toprule
Method & TF & PS & IR & PE & TG-Gen & TG-Edit & IM & Representation \\
\midrule
BlenderLLM~\cite{du2024blenderllm}
& $\times$ & $\times$ & $\times$ & $\times$
& \checkmark & $\times$ & $\times$
& CAD scripts (Blender Python/bpy) \\
LLaMA-Mesh~\cite{wang2024llama-mesh}
& $\times$ & $\times$ & $\times$ & $\times$
& \checkmark & $\times^{5}$ & $\times$
& Text-serialized mesh (vertices + faces) \\
NURBGen~\cite{usama2026nurbgen}
& $\times$ & $\times$ & $\times$ & $\times$
& \checkmark & $\times^{5}$ & $\times$
& NURBS patches and analytic primitives \\
CAD-Coder~\cite{doris2026cadcoder}
& $\times$ & $\times$ & $\times$ & $\times$
& $\times^{1}$ & $\times$ & \checkmark
& CadQuery (Python) \\
CADCodeVerify~\cite{al2025cadcodeverify}
& \checkmark & $\times$ & \checkmark & $\times$
& \checkmark & $\times$ & $\times$
& CAD scripting code (Python) \\
\midrule
ShapeCraft~\cite{zhang2026shapecraft}
& \checkmark$^{2}$ & \checkmark & \checkmark & \checkmark
& \checkmark & \checkmark & $\times$
& Graph-based Procedural Shape (GPS) \\
Articraft~\cite{zhou2026articraft} & \checkmark & \checkmark & \checkmark & \checkmark$^3$ & \checkmark & \checkmark$^3$ & \checkmark & CAD parts/joints (domain-specific SDK) \\
Prox-E~\cite{sella2026proxe}
& \checkmark & $\times^{4}$ & \checkmark & \checkmark
& $\times$ & \checkmark & $\times$
& Superquadric proxy \\
Shap-E~\cite{jun2023shap-e}
& $\times$ & $\times$ & $\times$ & $\times$
& \checkmark & $\times$ & \checkmark
& Implicit function parameters (mesh / NeRF) \\
CADCrafter~\cite{chen2025cadcrafter}
& $\times$ & $\times$ & $\times$ & $\times$
& $\times$ & $\times$ & \checkmark
& Parametric CAD command sequences \\
MvDrag3D~\cite{chen2024mvdrag3d}
& \checkmark$^{2}$ & $\times$ & \checkmark & \checkmark
& $\times$ & $\times$ & $\times$
& 3D Gaussians \\
\midrule
\textbf{SurfScaff3D (ours)}
& \checkmark & \checkmark & \checkmark & \checkmark
& \checkmark & \checkmark & \checkmark
& \textbf{Hierarchical B-spline patch scaffold (HPS)} \\
\bottomrule
\end{tabular}
\caption{
Comparison of capabilities and representations of related
3D generation and editing frameworks.
TF = training-free; PS = decomposition into explicitly labelled semantic
parts; IR = inference-time iterative/interactive refinement; PE =
part-level/local editing without regenerating the complete object;
TG-Gen = text-to-3D generation from scratch; TG-Edit = text-guided
editing of an existing asset; IM = image-conditioned input.
$^{1}$CAD-Coder's text input is a fixed instruction template rather than an object description, and all its reported results are image-conditioned.
$^{2}$Neither method trains a task-specific model, but both rely on per-instance test-time optimization (ShapeCraft uses score distillation for texturing, and MvDrag3D uses deformation and score distillation).
$^{3}$The Articraft paper describes only local edits made during generation, though its released code also supports text-guided editing of a completed asset.
$^{4}$Prox-E's superquadric primitives carry only a numeric index and colour rather than semantic part labels.
$^{5}$Neither method supports editing natively. For Table~3 of the main paper, we provide the source shape in each model's own format along with the edit instruction.
}
\label{tab:framework_comparison}
\end{table*}

\section{SurfScaff3D with Other Backbones}
We evaluate the SurfScaff3D framework with other open-weight backbones
on OmniObject3D (\autoref{tab:hps}). Semantic alignment remains within a narrow band across all three (CLIP 28.1--29.3, ULIP 22.8--26.1, VQA 60.5--64.8), and all three remain competitive on these metrics with the strongest baseline of Table~1 in the main paper (Articraft~\cite{zhou2026articraft}), indicating that the benefit of HPS is not tied to a single frontier model. Geometric fidelity is more sensitive to the choice of backbone, with FPD ranging from 384 (Qwen3.6) to 313 (DeepSeek-V4), though all three remain behind Articraft (234) and our Claude and Kimi configurations (206 and 211). Qwen3.6-35B-A3B~\cite{qwen2026qwen35b} operates with only 3B active parameters, less than a quarter of the other two backbones, yet the gap to the strongest backbone is limited to 1.2 CLIP and 4.3 VQA points, suggesting that HPS can remain effective at considerably smaller backbone scales.

\begin{table*}[h]
\centering
\scriptsize
\setlength{\tabcolsep}{4pt}
\begin{tabular}{l cc ccc c}
\toprule
Backbone & Total Params & Act.\ Params & CLIP$\uparrow$ & ULIP$\uparrow$ & VQA$\uparrow$ & FPD$\downarrow$ \\
\midrule
Qwen3.6-35B-A3B~\cite{qwen2026qwen35b}       & \phantom{0}35B & \phantom{0}3B & 28.1 & 22.8 & 60.5 & 384 \\
MiMo-V2.5~\cite{xiaomi2026mimov25}           & 310B & 15B & 29.1 & \textbf{26.1} & 62.4 & 344 \\
DeepSeek-V4-flash-0731~\cite{deepseek2026v4}  & 284B & 13B & \textbf{29.3} & 25.9 & \textbf{64.8} & \textbf{313} \\
\bottomrule
\end{tabular}
\caption{\textbf{SurfScaff3D with other backbones.} SurfScaff3D with each backbone, evaluated on OmniObject3D under the same protocol as Table 1 of the main paper. The MiMo and DeepSeek runs use three critic iterations, with Gemini 2.5 Pro~\cite{gemini25} as the critic. Active parameter counts are reported since all three
backbones are mixture-of-experts models.}
\label{tab:hps}
\end{table*}

\end{document}